%% file: main.tex
\documentclass[acmlarge]{acmart}
\AtBeginDocument{%
  }

\usepackage{amsmath, amssymb, amsfonts}
\usepackage{graphicx}
\usepackage{tabularx, booktabs, makecell, tabularray}
\usepackage{xspace}
\usepackage{xcolor}
\usepackage{enumitem}
\usepackage[caption=false, font=normalsize]{subfig}
\usepackage{algorithm}
\usepackage[noend]{algpseudocode}
\usepackage{lipsum}  
\usepackage{CJKutf8}
\usepackage{hyperref}
\usepackage{multirow}
\usepackage{appendix}
\usepackage{pgfplots}
\pgfplotsset{compat=1.18}

\newcommand{\ie}{\emph{i.e.} }

\newcommand{\eg}{\emph{e.g.,} }

\newcommand\figref[1]{Fig.~\ref{#1}}

\newcommand\tabref[1]{Tab.~\ref{#1}}

\newcommand\secref[1]{Sec.~\ref{#1}}

\newcommand{\fakeparagraph}[1]{\vspace{1mm}\noindent\textbf{#1.}}

\newcommand{\sysname}{{\sf UbiHearo}\xspace}
\newcommand{\sysnameposs}{{\sf UbiHearo's}\xspace}

\ifodd 1

\newcommand\lsc[1]{\textcolor{black}{#1}}

\newcommand\wfm[1]{\textcolor{black}{#1}}

\setcopyright{cc}
\setcctype{by}
\acmJournal{IMWUT}
\acmYear{2026} \acmVolume{10} \acmNumber{3} \acmArticle{170}
\acmMonth{9} \acmDOI{10.1145/3832010}

\begin{document}

\title{UbiHearo: Bringing Scenario-Aware Sound Guidance to DHH Users with Mobile Agents}

\author{Fengmin Wu}\authornote{Corresponding author}\affiliation{%
    \institution{Northwestern Polytechnical University}
    \country{China}
}\orcid{0009-0005-6965-623X}\email{fenny@mail.nwpu.edu.cn}

\author{Sicong Liu}\affiliation{%
    \institution{Northwestern Polytechnical University}
    \country{China}
}\orcid{0000-0003-4402-1260}\email{scliu@nwpu.edu.cn}

\author{Zimu Zhou}\affiliation{%
    \institution{City University of Hong Kong}
    \country{China}
}\orcid{0000-0002-5457-6967}\email{zimuzhou@cityu.edu.hk}

\author{Chenren Xu}\affiliation{%
    \institution{Peking University}
    \country{China}
}\orcid{0000-0001-9171-2596}\email{chenrenpku.edu.cn}

\author{Wanbing Zhao}\affiliation{%
    \institution{Northwestern Polytechnical University}
    \country{China}
}\orcid{0009-0005-0787-7328}\email{zhaowanbing@mail.nwpu.edu.cn}

\author{Peiwen Sun}\affiliation{%
    \institution{Northwestern Polytechnical University}
    \country{China}
}\orcid{0009-0003-8311-6198}\email{3130812498@mail.nwpu.edu.com}

\author{Zhiwen Yu}\affiliation{%
    \institution{Harbin Engineering University}
    \country{China}
}\orcid{0000-0002-9905-3238}\email{zhiwenyu@nwpu.edu.cn}

\renewcommand{\shortauthors}{Wu et al.}

\begin{abstract}
For Deaf and Hard-of-Hearing (DHH) users, effective sound awareness goes beyond detecting \emph{what} sound occurs; it requires understanding \emph{what it means}, \emph{how urgent it is}, and \emph{what to do}.
However, existing mobile assistants largely reduce acoustic scenes to isolated labels or generic alerts, failing to provide contextual guidance when the same sound implies different risks across scenarios.
We present \sysname, a mobile hearing assistant agent that closes the sensing--reasoning--action loop for \textit{scenario-aware} and \textit{personalized} DHH assistance.
\sysname adopts a \textit{neuro-symbolic} design that transforms continuous acoustic and physical context streams into structured intermediate representations (IRs) for controllable reasoning and guidance.
It introduces three key mechanisms.
\textit{First}, a perception-thresholded sensing mechanism enables energy-efficient always-on awareness by activating processing only for perceptually meaningful events while preserving safety-critical cues through deterministic overrides.
\textit{Second}, a staged neuro-symbolic reasoning framework performs promptless scenario understanding by conditioning an on-device audio-language model on structured IRs instead of user-authored context descriptions.
\textit{Third}, a bounded human-in-the-loop personalization mechanism adapts user-specific guidance through lightweight on-device decision-boundary updates while preserving population-aligned reasoning.
We evaluate \sysname on smartphones and in-vehicle systems across diverse scenarios.
It achieves favorable accuracy--latency--energy tradeoffs and personalized guidance.
\end{abstract}



\makeatletter

\let\UbiHearoOriginalAuthorsAddresses\@mkauthorsaddresses

\renewcommand{\@mkauthorsaddresses}{%
  \begingroup
    \def\@authornotemark{\textsuperscript{*}(corresponding author)}%
    \UbiHearoOriginalAuthorsAddresses
  \endgroup
  \unskip.\space
  \textsuperscript{*}For correspondence, contact Sicong Liu, Zhiwen Yu%
}

\renewcommand{\@authornotes}{}

\makeatother

\maketitle

\input{body/intro}

\input{body/res}

\input{body/sys}
\input{body/eval}
\input{body/rel}
\input{body/con}

\section*{Acknowledgments}
This work was partially supported by the National Natural Science Foundation of China (62522215, 62532009, 62472354, U25B2040, 62272010). 
Zimu Zhou's research is supported by CityU APRC grant (No. 9610633).


\bibliographystyle{ACM-Reference-Format}
\bibliography{main}
\newpage
\begin{appendices}
\input{body/appendix}
\end{appendices}
\end{document}
\vspace

%% file: body/intro.tex
\section{INTRODUCTION}

Deaf and Hard-of-Hearing (DHH) users face significant challenges because many \textit{safety-critical} and \textit{socially meaningful} sounds are inaccessible to them. 
Missing a door knock, alarm, or shouted warning can jeopardize safety, while missing speech cues in shared spaces undermines social interaction. 
While hearing aids and cochlear implants help restore auditory perception, they do not reliably provide \textbf{situational sound awareness} in dynamic mobile environments that vary across users, time, and locations. 
As a ubiquitous platform, smartphones provide built-in microphones and computing capabilities to continuously sense in-situ environmental sounds and deliver timely alerts, complementing hearing aids and serving as a fallback when they are unavailable.
This has driven the development of mobile sound-awareness apps~\cite{bib:ubiear,bib:Environmental,bib:AdaptiveSound,bib:Protosound,bib:SPECTRA}.

Despite advances, existing solutions still fall short of meeting DHH users' real needs in three critical areas, highlighted by our survey of 100 DHH users aged 15-60 (\secref{sec:background}):
\begin{enumerate}
    \item 
    \textbf{Sensing}.
    Many systems achieve high average recognition accuracy but still miss exactly the events DHH users care about most: \textit{brief}, \textit{low-SNR}, and \emph{overlapping events} that occur \textit{unpredictably} in daily life, \eg a shouted warning under traffic noise, a name call during a crowded commute, or an alarm mixed with TV/audio. 
    This is exacerbated by energy-saving duty-cycled sensing, which can \emph{skip} short transient cues.
    
    \item \textbf{Reasoning}.
    Most existing apps~\cite{bib:ubiear, bib:AdaptiveSound} map mixed physical scenes into isolated sound labels, \eg \emph{car horn}. 
    For DHH users, however, \emph{the label is not the answer}: the same sound can indicate very different meaning, urgency, and required attention depending on \textit{where they are}, \textit{what they are doing}, and \textit{how close/fast the source is}. 
    For instance, a horn heard indoors may be irrelevant, while a horn during street crossing demands immediate action (\figref{fig:case_intro}). 
    Without scenario-aware grounding, label-only alerts often lead to \emph{false urgency} (unnecessary interruptions) or \emph{missed urgency} (delayed reaction).

    \item \textbf{Guidance (Action)}.
    Label-only alerts shift interpretation burden back to DHH users, who must infer appropriate actions under time pressure and limited attention.
    Our survey shows strong demand for actionable situational guidance (\eg "check the door," "pause and look for traffic"), yet such assistance remains largely unexplored in existing mobile sound awareness systems.
\end{enumerate}
These observations motivate a mobile hearing assistant that operates as a \textbf{sensing-reasoning-action loop}.
It should continuously sense acoustic events under mobile energy budgets, interpret their situational meanings in \textit{user-specific physical contexts}, and provide timely guidance aligned with individual needs.

However, achieving this goal faces three fundamental challenges:

\begin{figure}[t]
    \centering 
    \vspace{-5mm}
    \subfloat[Indoor case]{\label{fig:acc}
    \includegraphics[width=0.26\linewidth]{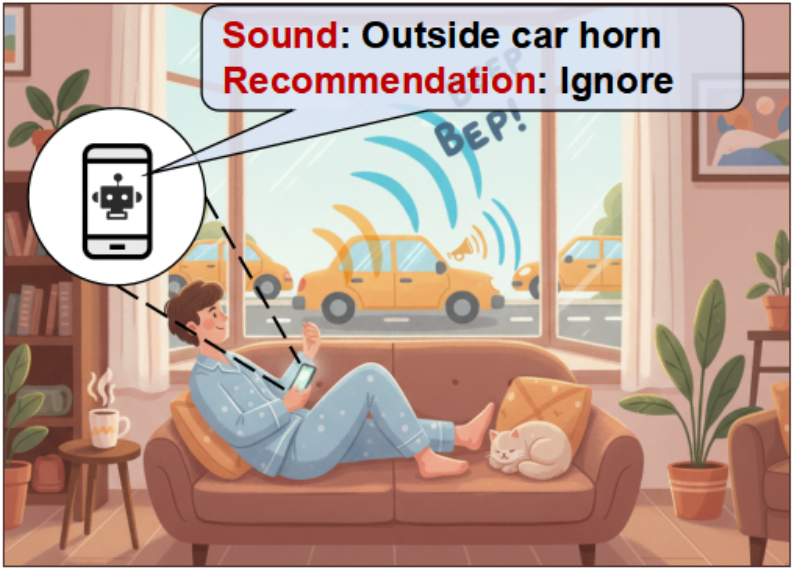}}
    \hspace{2mm}
     \subfloat[Outdoor case, street crossing]{\label{fig:laten}
    \includegraphics[width=0.26\linewidth]{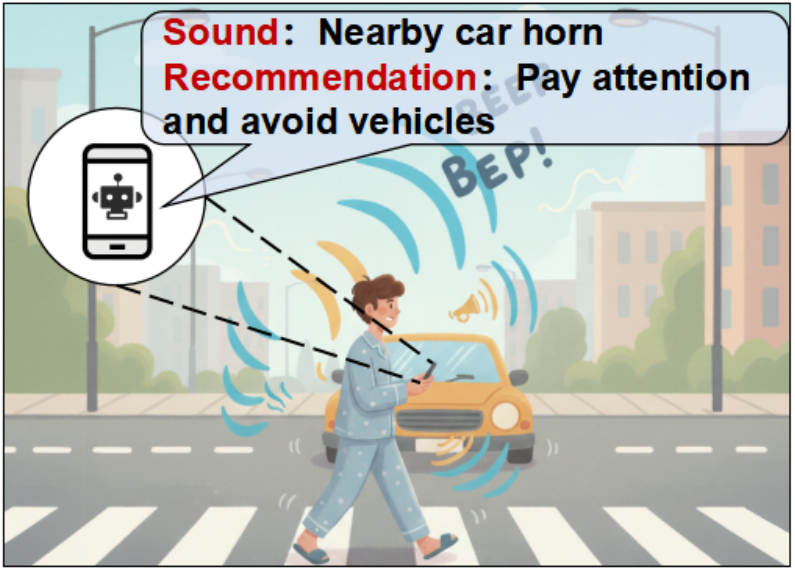}}
    \vspace{-3mm}
    \caption{Motivation cases, which illustrate scenario-dependent sound awareness for a DHH user, \textit{Lucas}.}
     \vspace{-2mm}
\label{fig:case_intro}
\end{figure}

\noindent$\bullet$ \textbf{Challenge \#1:}
Always-on awareness requires a difficult trade-off between recall and energy.
Continuous sensing provides high recall for critical events but incurs substantial energy overhead~\cite{bib:AdaptiveSound}, while aggressive duty cycling reduces energy consumption at the risk of missing brief and safety-critical sounds~\cite{bib:vad}.
The challenge is determining \textit{when and what to sense} to maintain continuous awareness under mobile energy constraints.

\noindent$\bullet$ \textbf{Challenge \#2:}
Sound awareness must answer not only \textit{what sound is detected}, but also \textit{what does it mean} \emph{here}, and \textit{what should I do now}, because physical context determines urgency and appropriate actions.
Although LLMs (\eg GPT~\cite{bib:gpt}, Qwen~\cite{bib:qwenb}) provide strong reasoning capabilities, their performance depends heavily on \textit{informative prompts} describing such context.
However, DHH users in mobile scenarios often cannot provide fine-grained prompts due to time pressure, limited attention and environmental awareness.
The challenge is enabling reliable scenario grounding and efficient reasoning without requiring user-authored prompts.

\noindent$\bullet$ \textbf{Challenge \#3.}
Scenario-aware reasoning captures population-level needs, but effective assistance is user-specific.
Users may differ in preferred sound priorities, interruption tolerance, and such preferences may evolve over time.
However, on-device personalization suffers from sparse and context-skewed feedback.
Naive updates may overfit individual corrections, drift from population- and scenario-aligned knowledge, and destabilize safety-critical guidance.
Existing DHH personalization approaches~\cite{bib:AdaptiveSound,bib:Protosound,bib:SPECTRA} mainly focus on sound-type customization and do not support scenario-conditioned preference adaptation.

To address these, we present \sysname, a mobile hearing assistant agent that provides \textit{scenario-aware} and \textit{personalized} sound awareness and guidance for DHH users.
It includes three key mechanisms.

\noindent$\bullet$ \textbf{First.}
We observe that human perception provides a natural sensing boundary: always-on awareness does not require always-on processing, as only perceptually meaningful events matter.
Based on this insight, \sysname introduces \textit{perception-driven sensing}, leveraging perceptual thresholds in spectrogram energy space to \textit{filter} irrelevant audio via short-term thresholded listening, \textit{extract} compact acoustic IRs via perceptual frequency decomposition and lightweight GMM modeling, and \textit{activate} context sensing only upon meaningful events.

\noindent$\bullet$ \textbf{Second.}
We observe that scenario-aware assistance requires hierarchical reasoning: grounding the user's physical context \textit{before} interpreting sound meaning.
Moreover, LLMs require informative prompts \textit{mainly because} contextual states are not explicitly available during inference.
Thus, \sysname introduces a \textit{neuro-symbolic staged reasoning} mechanism.
It first adopts a 1.5B-parameter Qwen2.5-Instruct  model to infer physical context, then transforms heterogeneous context cues and acoustic tags into symbolic scenario states (\ie IRs), which serve as \textit{structured prompts} for an on-device 1.8B-parameter Qwen2-Audio model to generate scenario-aware guidance.

\noindent$\bullet$ \textbf{Third.}
We note that personalization should adapt \textit{how to respond} rather than \textit{what to understand}, since acoustic semantics and physical context reasoning are shared while guidance preferences are user-specific.
Harnessing this, \sysname updates a lightweight output-layer adapter from per-instance feedback, and leverages regularization to constrain updates as bounded deviations from population-aligned behavior, enabling private adaptation while reducing overfitting, preference drift, and online oscillation.

We prototype \sysname on smartphones and in-vehicle systems to deliver end-to-end sensing, scenario-aware notifications, and actionable guidance.
Across diverse datasets and real-world tests spanning varied acoustic conditions and physical contexts, \sysname consistently outperforms five state-of-the-art baselines.
It reduces miss rate by up to 10.3\% while consuming 391$\times$ less energy, enabling up to 12.5\,h of continuous operation under high-frequency trigger setting.
For reasoning, \sysname improves scenario-aware user preference alignment by up to 61\% via stage-wise task decomposition and scenario-conditioned preference training, while reducing model parameters by 53.8\%.
\sysname runs near real time on-device (\eg 0.9\,s on Google Pixel~9), whereas audio-MLLM baselines (\eg Qwen-Audio2~\cite{bib:qwen}, Audio-Flamingo~\cite{bib:flamingo}, MiDashengLM~\cite{bib:midashenglm}) remain too resource-intensive for on-device deployment.
In summary, we make the following contributions: 
\begin{itemize}
    \item 
    We identify and formalize a key requirement for DHH hearing assistance: moving beyond sound labeling to scenario-aware understanding and actionable physical guidance, grounded in a DHH user study.
    \item 
    We design \sysname, a neuro-symbolic mobile assistant that unifies sensing, reasoning, and action through perceptual-thresholded duty-cycled sensing, structured intermediate representations, route-conditioned on-device reasoning, and human-in-the-loop personalization.
    \item 
    We test \sysname on diverse mobile scenarios, showing that it improves scenario-aware reasoning, user-aligned guidance, and accuracy-latency-energy trade-offs, enabling practical near-real-time DHH assistance.
\end{itemize}

%% file: body/res.tex
\begin{figure}[t]
    \centering
    \vspace{-5mm}
    \subfloat[Missing rates]{\label{fig:q1}
        \includegraphics[width=0.25\linewidth]{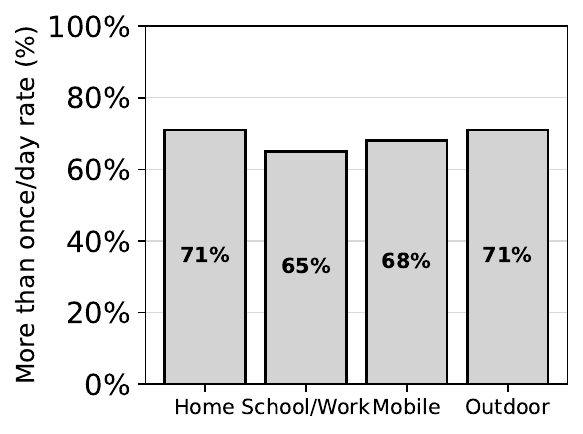}}
    \subfloat[Requiring guidance rates]{\label{fig:q2}
        \includegraphics[width=0.25\linewidth]{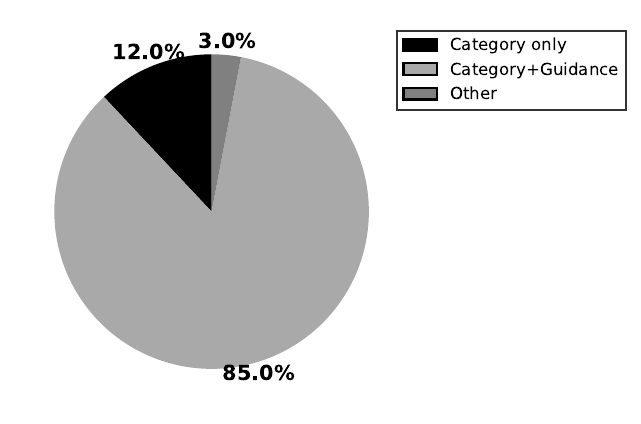}}
    \subfloat[Impact rates]{\label{fig:q3}
        \includegraphics[width=0.25\linewidth]{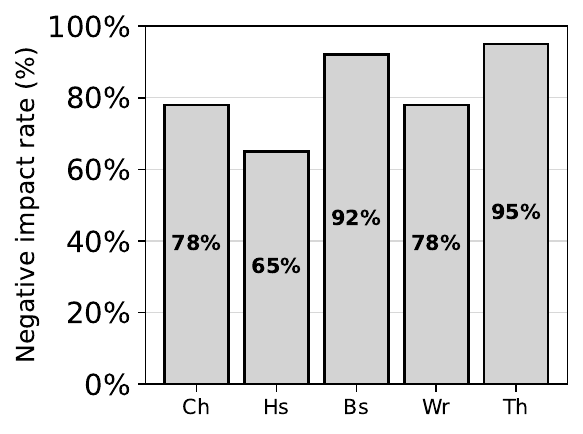}}
    \subfloat[Preference]{\label{fig:q4}
        \includegraphics[width=0.25\linewidth]{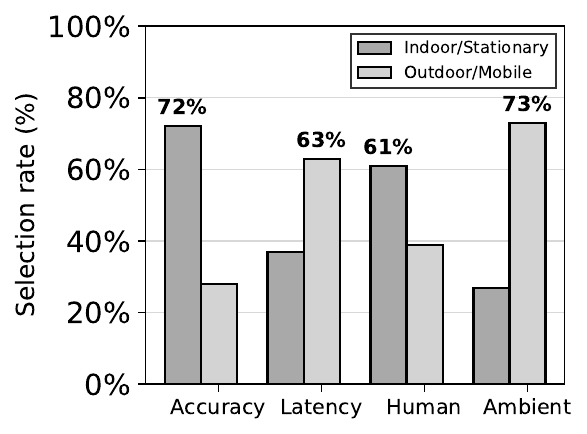}}
    \caption{Survey results. (a) Missing rates of acoustic events, (b) Expectations for guidance, (c) Impact of universal guidance, and (d) Scenario-specific preference.}
    \label{fig:1}
    \vspace{-5mm}
\end{figure}

\section{BACKGROUND AND MOTIVATION}
\label{sec:background}
To ground our design in real user needs, we conducted a questionnaire (Appendix \ref{questionnaire}) study with 100 DHH participants (aged 15$\sim$60), combining multiple-choice and open-ended questions (\figref{fig:1}).
Permission was obtained from the participants. 
The results reveal that effective mobile hearing assistance must go beyond sound labeling, and instead provide \textit{scenario-aware interpretation}, \textit{situational guidance}, and \textit{instant personalized trade-offs}.

Formally, we define \textbf{Scenario} as a semantically grounded, user-centric \textit{state} that links \textit{i)} the user’s location and activity, \textit{ii)} the spatial relationship between the sound source and the user,
and \textit{iii)} the surrounding acoustic context (\eg foreground/background mixing of speech and ambient sounds).
Scenario grounding is essential for translating detected sounds into appropriate user guidance, while dynamically controlling the system trade-off among accuracy, urgency, and resource efficiency under different contexts.

\subsection{Survey Findings}
\label{sec:background:survey}

\subsubsection{Frequently Missed Speech and Ambient Sound Events}
\label{subsec:missed}
The survey shows that 91\% of participants own a smartphone and are willing to use a mobile app for sound awareness, suggesting high acceptability.
Yet reliable sound awareness remains difficult: 71\%, 65\%, 68\%, and 71\% report missing acoustic events more than once per day at home, at school/work, while mobile, and outdoors, respectively (\figref{fig:q1}).
Participants attributed these misses to short-duration cues, mixed speech and ambient sounds, noisy environments, and overlapping sources.
These findings highlight the need for always-on mobile sensing that reliably captures context-critical events while maintaining energy efficiency.

\subsubsection{Expectations Beyond Label Alerts}
\label{sec:survey_guidance}
Participants need more than sound labels.
As shown in \figref{fig:q2}, 85\% expect \textit{situational guidance} after receiving alerts; 
among them, 79\% (67/85) find labels alone insufficient to judge event urgency or determine appropriate actions, and 81\% (69/85) believe situational guidance would improve safety.
Consistent with prior psychological findings~\cite{bib:pg}, participants prefer guidance such as ``check the door'' or ``stay cautious near traffic'' over isolated label (\eg ``door knock'').
Such guidance reduces cognitive burden when DHH users must make rapid decisions under limited sound awareness and attention.

\subsubsection{Scenario Awareness Determines Appropriate Guidance}
\label{sec:survey_scenario}
We further asked participants to compare \textit{scenario-aware guidance} with \textit{scenario-agnostic universal guidance} for five representative acoustic events (\tabref{tab:tab1}).
Because the same sound may imply different urgency and actions across scenarios, scenario-agnostic universal guidance can be ineffective or misleading.
As shown in \figref{fig:q3}, participants reported such issues for car horns (78\%), human shouting (65\%), broadcast sounds (92\%), running water (78\%), and thunder (95\%).
These highlight the \textit{essential role} of scenario-aware reasoning and guidance.

\begin{table}[t]
\vspace{-3mm}
\caption{Examples of guidance strategies in the Questionnaire for common acoustic events across various scenarios}
\tiny
\vspace{-2mm}
\label{tab:tab1}
\begin{tabular}{|c|c|c|c|}
\hline
Acoustic event & Scenario & Scenario-aware guidance & Assumed-universal guidance \\ \hline
\multirow{2}{*}{Car horn(Ch)} & Indoor & Ignore & \multirow{2}{*}{Ignore} \\ \cline{2-3}
& On the road & Attention to vehicle & \\ \hline
\multirow{2}{*}{Human shouting(Hs)} & Playground & Ignore & \multirow{2}{*}{Ignore} \\ \cline{2-3}
& Outdoor & Attention to source & \\ \hline
\multirow{2}{*}{Broadcast sound(Bs)} & At school/square & Ignore & \multirow{2}{*}{Ignore} \\ \cline{2-3}
& Railway station/Airport & Attention to departure time & \\ \hline
\multirow{2}{*}{Water running(Wr)} & Home & Turn off the tap & \multirow{2}{*}{Turn off the tap} \\ \cline{2-3}
& Wild & Attention to water source & \\ \hline
\multirow{2}{*}{Thunder(Th)} & Indoor & Ignore & \multirow{2}{*}{Ignore} \\ \cline{2-3}
& Outdoor & Attention to lighting & \\ \hline
\end{tabular}
\vspace{-3mm}
\end{table}


\subsubsection{Scenario-varying User Preferences}
\label{sec:3.1.3}
DHH users' preferences dynamically vary across scenarios (\figref{fig:q4}).
When indoors and stationary, 72\% prioritize higher accuracy, whereas outdoors and mobile, 63\% prioritize faster responsiveness.
Their event priorities also shift with physical context: 73\% emphasize safety-critical ambient sounds outdoors (\eg traffic), while 61\% emphasize human speech and socially meaningful cues indoors.
These show that user preferences depend on context and individual needs, making fixed configurations suboptimal.

\subsection{Motivating Example}
\label{moti_case}
Consider \textit{Lucas}, a DHH user who relies on a mobile assistant for sound awareness.
At home, \textit{Lucas} is stationary and prefers accurate recognition, while many non-human sounds are low priority.
When a car horn is heard outside the building (\figref{fig:acc}), the assistant should recognize the event, understand that it poses limited urgency in this indoor context, and avoid unnecessary interruptions.
Later, when \textit{Lucas} walks outdoors, responsiveness becomes critical because traffic sounds may indicate immediate safety risks.
If the same car horn occurs near a street crossing (\figref{fig:laten}), the assistant should detect it promptly and provide actionable guidance (\eg checking surroundings) while continuing to monitor subsequent cues.
This example reveals a fundamental requirement: \textbf{scenario context determines both the meaning of an acoustic event and the appropriate system response}. Therefore, practical mobile hearing assistance must adapt its sensing strategy and guidance according to changing scenarios.


\subsection{Design Goals}
\label{sec:goal}
Guided by the above findings, a mobile hearing assistant should satisfy three design goals:
\begin{itemize}
    \item \textbf{Energy-efficient always-on sensing.}
    Survey results (\secref{subsec:missed}) show that missed speech and ambient sound events pose safety risks. 
    The system should support continuous acoustic monitoring under strict battery constraints while preserving sensitivity to user-critical events.

    \item \textbf{Scenario-aware sensing-reasoning-action.} 
    Findings from \secref{sec:survey_guidance} and \secref{sec:survey_scenario} show that sound meaning and appropriate responses depend on scenario context.
    Instead of producing isolated labels, the system should integrate acoustic signals with physical context (\eg location, motion, and activity) to infer scenarios, reason about urgency, and provide actionable guidance.

    \item \textbf{On-device human-in-the-loop personalization.}
    DHH users’ preferences are highly individual, scenario-dependent, and evolve over time (\secref{sec:3.1.3}), making static models insufficient. 
    The system should support on-device, human-in-the-loop personalization, enabling incremental adaptation via user feedback and lightweight fine-tuning without cloud retraining or privacy risks.

\end{itemize}

%% file: body/sys.tex
\begin{figure}[t]
    \centering
    \vspace{-2mm}
\includegraphics[width=0.98\linewidth]{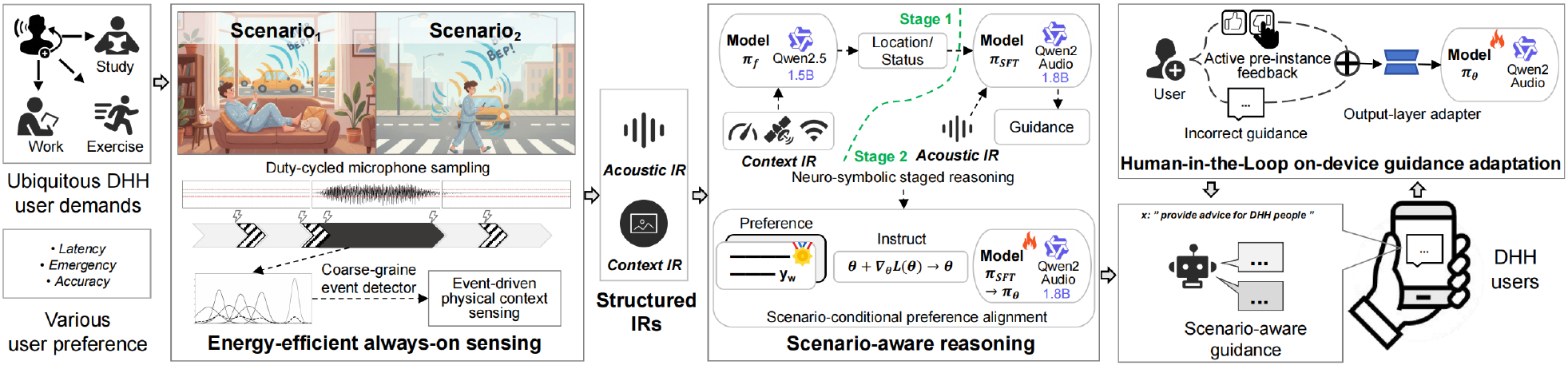}
    \caption{Workflow of \sysname.}
    \label{fig:fw}
    \vspace{-2mm}
\end{figure}

\section{SYSTEM DESIGN}

\subsection{Overview}
We design \sysname as a mobile hearing assistant that transforms continuous acoustic and contextual signals into scenario-aware notifications and actionable guidance through a \textbf{sensing--reasoning--action pipeline}.
\sysname adopts a \textbf{neuro-symbolic design}~\cite{bib:neuro}, where lightweight sensing modules extract reliable evidence from the environment, intermediate representations (IRs) provide symbolic states for scenario grounding, and an aligned audio-language model generates user-facing guidance.
This design moves beyond sound classification and black-box multimodal reasoning by enabling controllable scenario-level interpretation.

\begin{itemize}
    \item \textbf{Sensing.}
    To enable energy-efficient always-on awareness, \sysname employs a perception-thresholded sensing mechanism inspired by the \textit{Weber--Fechner law}~\cite{bib:wfl}.
    The key insight is that always-on awareness does not require always-on processing:
    \sysname performs lightweight \textit{short-term listening} to suppress silence and irrelevant noise, and activates \textit{long-term listening} only when perceptually meaningful events emerge; otherwise, it enters \textit{sleep} to reduce idle energy.
    Rather than making final decisions, this module extracts reliable acoustic evidence from continuous audio streams through perceptually motivated frequency analysis and a lightweight GMM front end.
    Verified safety-critical events bypass reasoning and directly trigger deterministic alerts  (\secref{sec:4.1}).

    \item \textbf{Structured Intermediate Representation (IR)}. 
    Raw sensing outputs alone are insufficient for scenario understanding because acoustic meaning depends on physical context.
    \sysname therefore converts sensing outputs into two IRs:
    an \textit{acoustic IR} describing event evidence and safety relevance, and a \textit{context IR} describing physical states such as mobility and indoor/outdoor status.
    These IRs provide a symbolic interface between sensing and reasoning, enabling automatic scenario grounding without requiring DHH users to provide context descriptions under time pressure and limited attention.

    \item \textbf{Reasoning}. 
    \sysname reformulates DHH scenario reasoning from end-to-end black-box inference into a \textit{neuro-symbolic staged reasoning process}.
    Instead of directly mapping raw multimodal inputs to responses, \sysname first infers physical scenarios from lightweight sensor cues and then reasons about acoustic events conditioned on structured IRs.
    The resulting IR-guided prompts enable an on-device audio-language model to generate scenario-aware notifications and guidance while reducing multimodal ambiguity and supporting scenario-conditioned preference alignment (\secref{sec:reason}).

    \item \textbf{Action (guidance).}
    \sysname maps inferred scenarios into concise guidance (\eg \textit{ignore}, \textit{investigate}, or \textit{act cautiously}).
    Beyond static rules or cloud-based retraining, \sysname enables \textit{human-in-the-loop on-device adaptation}: user feedback on incorrect interpretations or guidance is locally converted into user-specific supervision signals.
    Instead of updating the entire model, \sysname adapts only a lightweight output-layer adapter, localizing personalization at the final decision boundary while preserving general acoustic and scenario reasoning (\secref{sec:action}).

\end{itemize}

\begin{figure}[t]
    \centering 
    \vspace{-2mm}
    \includegraphics[width=0.6\linewidth]{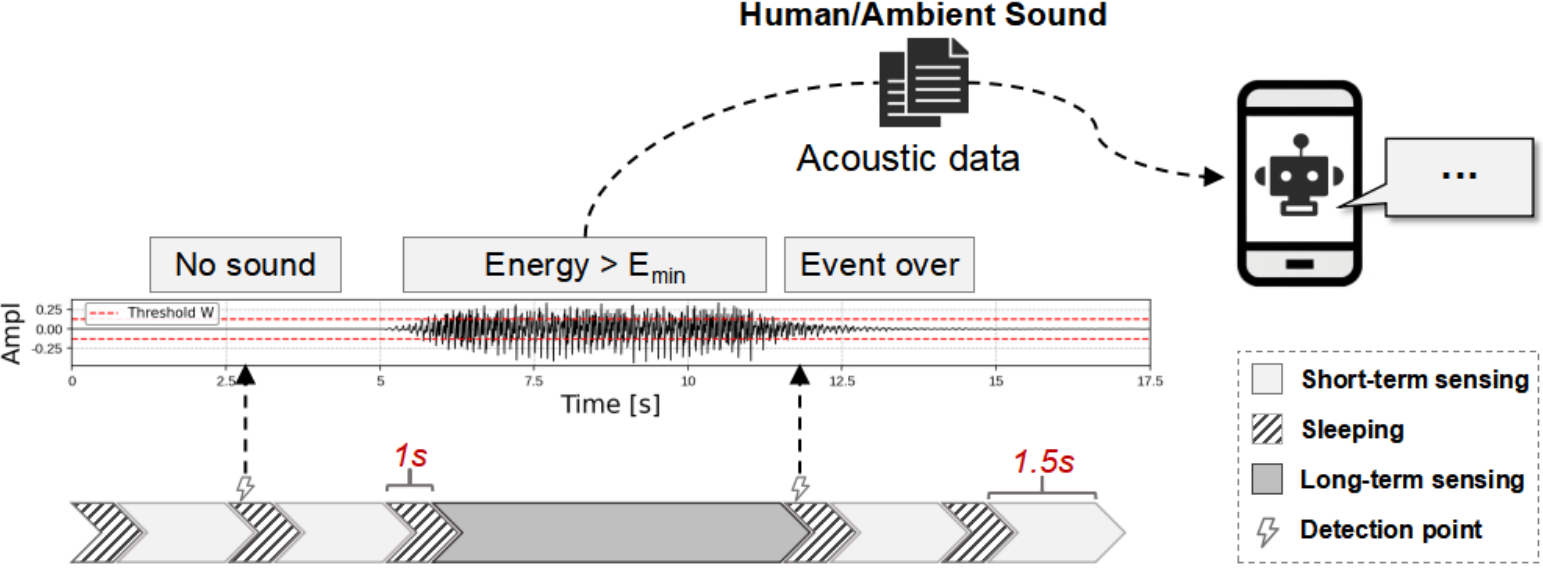}
    \caption{Illustration of the duty-cycled microphone sampling mechanism.}
    \label{fig:duty}
    \vspace{-5mm}
\end{figure}

\subsection{Energy-Efficient Always-On Sensing}
\label{sec:4.1}
\subsubsection{Duty-Cycled Microphone Sampling}
\label{sec:4.1.1}
DHH users require continuous sound awareness, yet most environmental audio contains no meaningful information.
Existing approaches face a recall--energy dilemma:
continuous microphone processing provides reliable detection but consumes excessive power~\cite{bib:AdaptiveSound}, whereas fixed or coarse-grained adaptive sampling may miss short-lived alarms, sirens, or verbal calls~\cite{bib:SPECTRA,bib:Protosound}.
\sysname resolves this by grounding sensing decisions in \textit{perceptual detectability} rather than fixed sampling schedules.
Our key observation is that sound perception has an intrinsic energy boundary:
acoustic events below the just-perceptible energy threshold ($E_{\min}\approx10^{-6}$~J) are fundamentally inaudible to humans~\cite{bib:jnd}.
Therefore, sensing should preserve events that accumulate sufficient perceptual energy instead of continuously processing all audio.
For a sound with duration $T$ and perceived bandwidth $B$, its accumulated energy can be approximated as:
\begin{equation}
    E \propto \sum_{n} \int_{B}|x(n, f)|^{2} d f \cdot \Delta t,
\end{equation}
where $|x(n,f)|^{2}$ denotes the energy density and $\Delta t$ represents the temporal integration window.
This formulation provides a perceptual basis for mapping human sound perception into sensing decisions.

Based on this principle, \sysname adopts a duty-cycled listening mechanism.
Because DHH-relevant events are temporally sparse, \sysname first performs \textit{short-term listening} for inexpensive event discovery.
During this stage, it suppresses silence and background noise and produces \textit{coarse} acoustic cues.
When a non-silent event is detected, \sysname switches to \textit{long-term listening} to capture richer acoustic information; 
otherwise, it enters \textit{sleep} mode to reduce idle energy.
Importantly, the sensing layer does not make final semantic decisions.
Instead, it filters irrelevant audio and converts continuous streams into compact cues for downstream scenario reasoning.
\sysname maps perceptual constraints into sensing parameters as follows: 
\begin{itemize} 
\item \textit{Sampling rate.}
\sysname uses 32~kHz sampling to preserve spectral information within the main human-audible frequency range and support subsequent event verification.
\item \textit{Frame length.}
\sysname adopts 20~ms frames as the energy integration window, balancing stable short-time energy estimation and low detection latency.
\item \textit{Duty-cycled schedule.}
\sysname samples audio in 10-s cycles at 0, 2.5, 5, and 7.5~s.
Intervals that fail to accumulate sufficient energy toward $E_{\min}$ are skipped, reducing unnecessary sensing with minimal perceptual loss.
\item \textit{Short-term listening duration.}
\sysname dynamically adjusts short-term listening duration according to battery level:
[0.02--1.60~s] when battery $>50\%$ and [0.02--1.00~s] otherwise.
These bounds remain above perceptual integration requirements while limiting sensing overhead.
\end{itemize}

\begin{figure}[t]
    \centering
    \vspace{-5mm}
    \subfloat[Seven non-uniform frequency sub-bands]{\label{fig:gmm1}
        \hspace{-5mm}\includegraphics[height=0.23\linewidth]{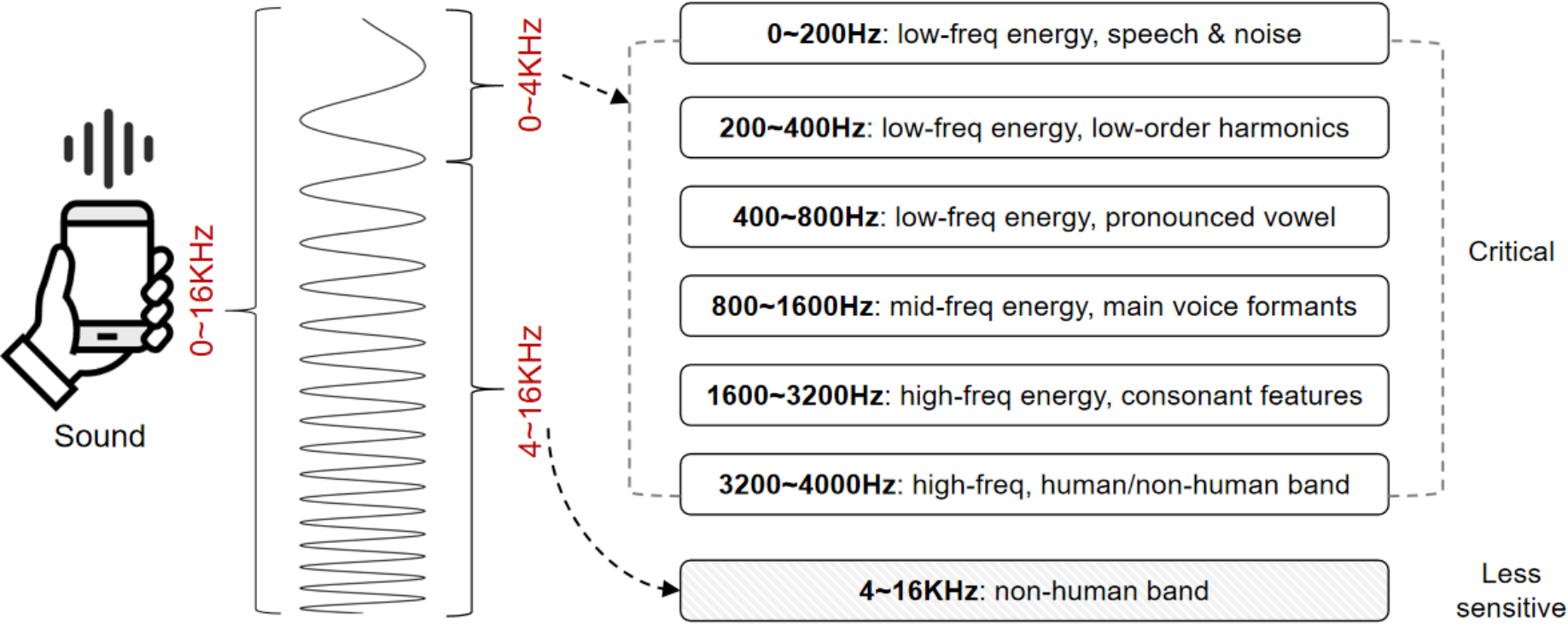}}
    \subfloat[The spectrogram of human talk and car horn]{\label{fig:gmm2}
        \includegraphics[height=0.23\linewidth]{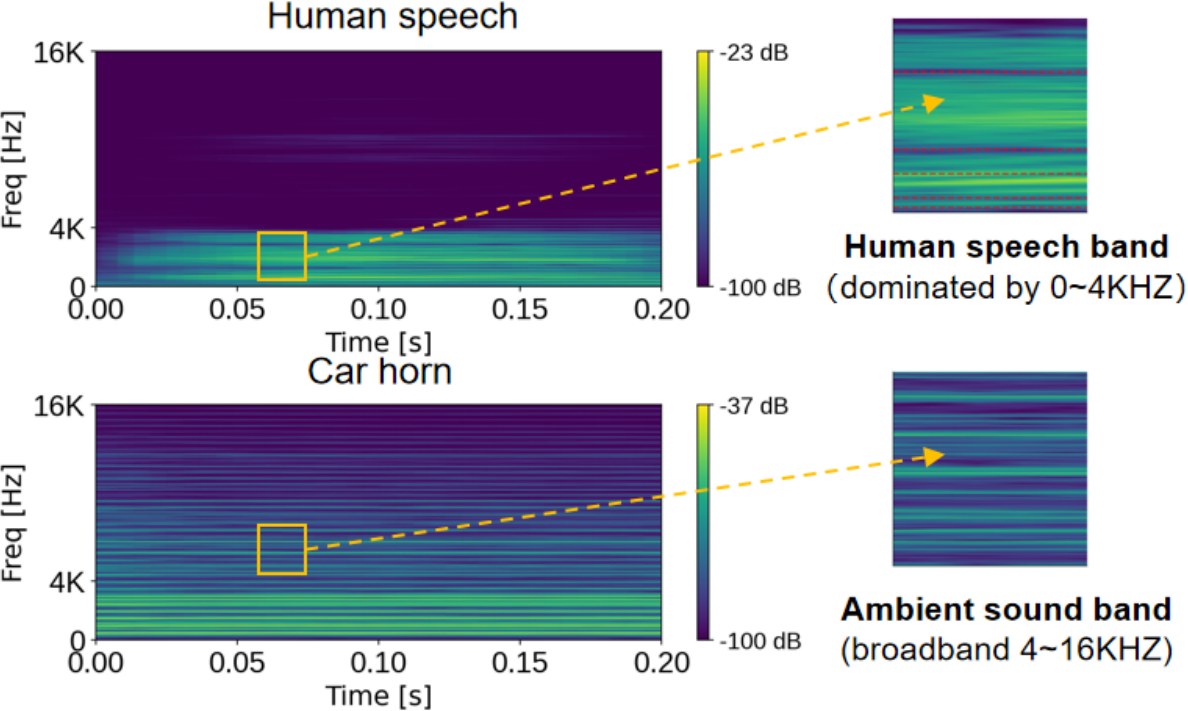}}
    \caption{The basis for non-uniform frequency band division.}
    \vspace{-5mm}
    \label{fig:gmm}
\end{figure}

\subsubsection{Coarse-grained Acoustic Event Detector}
\label{sec:4.1.2}
After perceptual-thresholded sensing identifies potentially meaningful audio, \sysname extracts compact acoustic evidence rather than performing expensive semantic classification.
The goal is not to determine final sound meaning, but to provide reliable and interpretable acoustic cues for downstream scenario reasoning.
Therefore, \sysname employs a lightweight GMM detector instead of neural sound classifiers (\eg YAMNet~\cite{bib:yamnet}, Silero~\cite{bib:silero}).
The detector exploits an acoustic asymmetry between speech and ambient sounds:
speech exhibits structured energy patterns in low- and mid-frequency bands, whereas ambient sounds have more diverse spectral distributions.
Accordingly, \sysname divides each frame into seven non-uniform frequency sub-bands (\figref{fig:gmm1}), assigning finer resolution to lower bands where speech-related cues are more perceptually salient under the \textit{Weber-Fechner law}~\cite{bib:wfl}.
As shown in \figref{fig:gmm2}, speech energy is mainly concentrated in 0--4~kHz, while ambient sounds spread more broadly across 0--16~kHz.
For each frame $t$, \sysname computes the energy of sub-band $i$ as:
\begin{equation}
E_{t,i}=\int_{B_i}|X_t(f)|^2 df, \quad
x_{t,i}=\log(E_{t,i}+\epsilon),
\end{equation}
where $B_i$ denotes the $i$-th sub-band, $X_t(f)$ represents the short-time spectrum, and $\epsilon$ ensures numerical stability.
\sysname models each sub-band log-energy distribution using a GMM:
\begin{equation}
p_i(x_{t,i})=\sum_{j=1}^{K_i} w_{i,j}\mathcal{N}
\left(x_{t,i}\mid \mu_{i,j},\sigma_{i,j}^{2}\right),
\end{equation}
where $w_{i,j}$, $\mu_{i,j}$, and $\sigma_{i,j}^{2}$ denote the mixture weight, mean, and variance of the $j$-th component.
Since the GMM operates in log-energy space, the expected physical energy is estimated as:
\begin{equation}
\bar{E}_i
= \mathbb{E}_{p_i}\left[\exp(x)\right]
= \sum_{j=1}^{K_i} w_{i,j}
\exp\left(\mu_{i,j}+\frac{1}{2}\sigma_{i,j}^{2}\right).
\end{equation}

During online detection, \sysname first checks whether observed energy exceeds the perceptual threshold $\tau_E$ derived from $E_{\min}$:
\begin{equation}
E_t=\sum_{i=1}^{M}E_{t,i}
= \sum_{i=1}^{M}\int_{B_i}|X_t(f)|^2df.
\end{equation}
where $M$ is the number of sub-bands.
For detected events, \sysname uses GMM likelihoods $\{p_i(x_{t,i})\}_{i=1}^{M}$ to estimate speech consistency.
High-energy frames with low speech consistency are conservatively classified as ambient sounds, reducing false speech detections.
The detector outputs an \textit{acoustic IR} containing event type, confidence score, and safety relevance, which serves as the interface to scenario reasoning (\secref{sec:reason}).
For safety-critical sounds, \sysname adds a deterministic verification layer using band-wise energy ratios and temporal consistency (Appendix~\ref{algorithm}).
Verified alarms and sirens bypass LLM reasoning and directly trigger alerts; uncertain events receive conservative notifications.

\subsubsection{Event-Driven Physical Context Sensing}
Scenario reasoning also requires physical context, but continuously sensing all modalities wastes energy because environmental context evolves more \textit{slowly} than acoustic events.
Thus, \sysname adopts event-driven context acquisition: it activates physical sensing only after detecting perceptually meaningful acoustic events.
Upon an acoustic trigger, \sysname extracts three lightweight context cues:
\textit{step frequency} for mobility estimation,
\textit{Wi-Fi density and SSID semantics} for indoor context inference,
and \textit{satellite visibility with SNR} for outdoor confirmation.
These robust physical indicators~\cite{bib:penetrative} are mapped into structured context labels, forming a \textit{context IR}.
Together, the acoustic IR and context IR provide automatically generated structured inputs for downstream scenario reasoning.

\subsection{Scenario-Aware Reasoning}
\label{sec:reason}

Effective DHH assistance requires grounding acoustic evidence into user-specific scenarios and generating appropriate responses.
However, directly applying generic audio-language models to mobile DHH assistance is insufficient.
\textit{First}, large multimodal models are difficult to deploy under mobile latency and energy constraints.
\textit{Second}, feeding raw audio and heterogeneous sensor streams directly into a generic model entangles perception and reasoning, increasing inference cost and hallucination risks.
\textit{Third}, models trained on general hearing-population data lack awareness of DHH users' scenario-dependent priorities, such as when to ignore, investigate, or immediately react to an event.
\sysname addresses these through an IR-guided two-stage reasoning architecture.
The key \textit{insight} is that scenario reasoning should be explicitly grounded before acoustic interpretation.

\subsubsection{Neuro-symbolic Staged On-Device Reasoning}
\label{sec:pl}
\sysname follows two \textit{observations} to design its reasoning pipeline.
\textit{First}, physical context inference and acoustic interpretation represent different reasoning dimensions.
Physical states, such as mobility and indoor/outdoor status, are relatively low-dimensional and can be inferred from lightweight sensors, whereas acoustic meaning requires richer semantic reasoning.
Separating these two processes avoids forcing a single model to jointly learn heterogeneous modalities and reduces unnecessary reasoning complexity.
\textit{Second}, compact models become effective when their inputs are structured and aligned with user preferences.
Prior work shows that preference alignment can enable smaller models to outperform larger generic models in human evaluations~\cite{bib:Traininglanguage}.
Therefore, instead of directly prompting a large model with raw observations, \sysname provides structured IRs that encode scenario-relevant information.

\begin{figure}[t]
    \centering
    \vspace{-5mm}
    \subfloat[Scheme 1]{\label{fig:cs2}
        \includegraphics[height=0.12\linewidth]{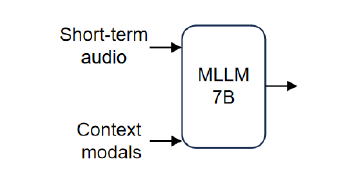}\label{scheme1}}
    \subfloat[Scheme 2]{\label{fig:cs3}
        \includegraphics[height=0.12\linewidth]{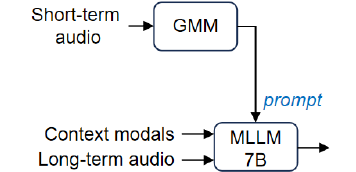}\label{scheme2}}
    \subfloat[UbiHearo (Two-stage)]{\label{fig:csr}
        \includegraphics[height=0.12\linewidth]{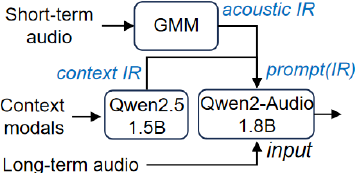}\label{scheme3}}
    \caption{Comparison of three reasoning schemes. (a) A single MLLM directly processes all modalities. (b) A 7B MLLM reasons over GMM-derived prompts. (c) \sysname uses two lightweight models, combining contextual states and GMM outputs as structured prompts.}
    \vspace{-5mm}
    \label{fig:cs}
\end{figure}

Based on these observations, \sysname adopts a two-stage IR-guided on-device reasoning pipeline (\figref{scheme3}), where sensing-generated IRs automatically replace manual user prompts.

\begin{itemize} 
\item \textit{Stage~1: Physical context inference.} 
\sysname first infers structured physical states from lightweight sensor cues.
A compact modality-fusion model based on Qwen2.5-1.5B-Instruct processes accelerometer signals, Wi-Fi features, and satellite visibility to infer context labels, such as \textit{indoor/outdoor} and \textit{stationary/mobile}.
These labels form an explicit scenario scaffold for downstream Stage~2.


\item \textit{Stage~2: Scenario reasoning and guidance generation.}
Given the inferred context IR and acoustic IR, \sysname constructs structured prompts for an on-device audio-language model.
The model is derived from Qwen2-Audio-7B-Instruct and compressed to 1.8B parameters through hierarchical quantization~\cite{bib:quantization} and low-rank decomposition~\cite{bib:lora}.
It jointly reasons over acoustic evidence, physical context, and acoustic tags to generate scenario-aware notifications and actionable guidance.

\end{itemize}


\textit{Why Structured IRs Help.}
From a \textit{neuro-symbolic} perspective, IRs serve as explicit scenario states rather than hand-crafted rules.
They bridge low-level perception and high-level reasoning by transforming continuous multimodal observations into structured, low-entropy inputs.
Moreover, LLM reasoning heavily depends on informative prompts, which are difficult for DHH users to provide during mobile interaction due to limited environmental awareness, attention, and time pressure.
\sysname therefore automatically constructs \textit{structured prompts} through IRs, organizing user situations into scenario routes such as \textit{indoor/stationary}, \textit{indoor/mobile}, \textit{outdoor/stationary}, and \textit{outdoor/mobile}.
These routes guide reasoning and align responses with scenario-specific priorities.
Thus, \sysname enables reliable scenario interpretation without requiring users to manually describe their context, while allowing a general audio-language model to perform scenario-specialized reasoning.


\subsubsection{Training Strategy}
Scenario-aware reasoning requires both stable scenario grounding and alignment with DHH users' preferences.
However, directly optimizing an end-to-end model may blur the boundary between context inference, acoustic interpretation, and guidance generation, while generic preference optimization cannot capture scenario-dependent needs.
Therefore, \sysname adopts a \textit{two-step} training strategy:
\textit{stage-wise supervised fine-tuning} (SFT) establishes functional interfaces between reasoning stages, and \textit{scenario-conditioned direct preference optimization} (DPO) aligns final guidance with DHH-specific preferences.

\begin{figure}[t]
    \centering
    \vspace{-5mm}
    \subfloat[SFT process of modal-fusion model \(\pi_{f}\) ]{\label{fig:sft1}
        \includegraphics[height=0.26\linewidth]{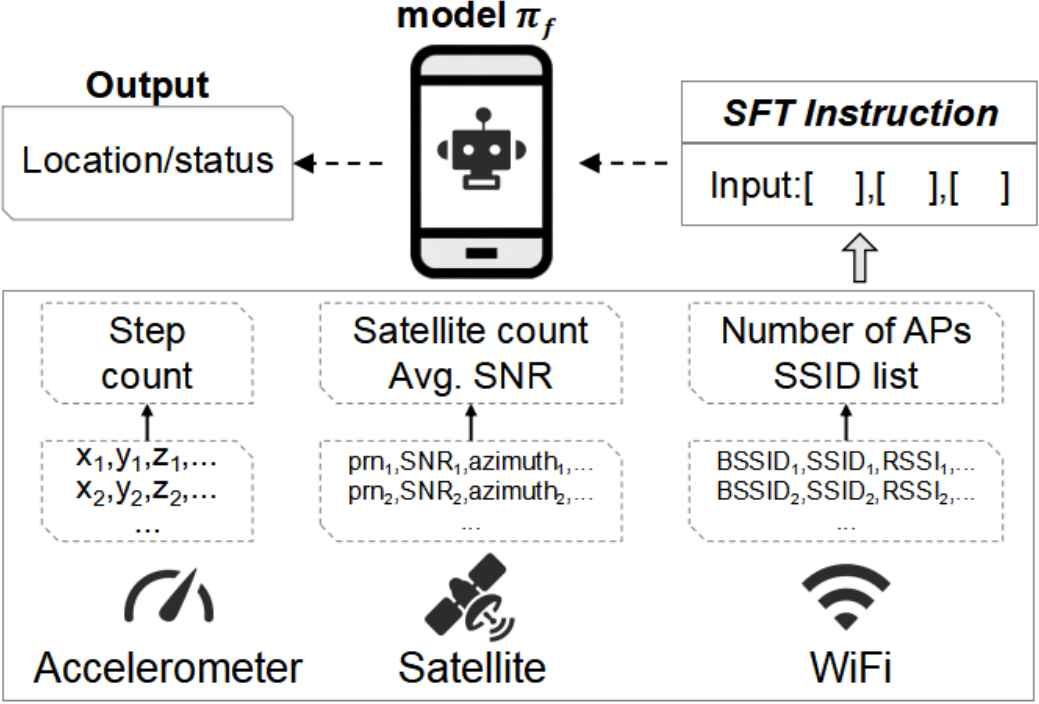}}
    \hspace{5mm}\subfloat[SFT process of scenario reasoning model \(\pi_{SFT}\)]{\label{fig:sft2}
        \includegraphics[height=0.26\linewidth]{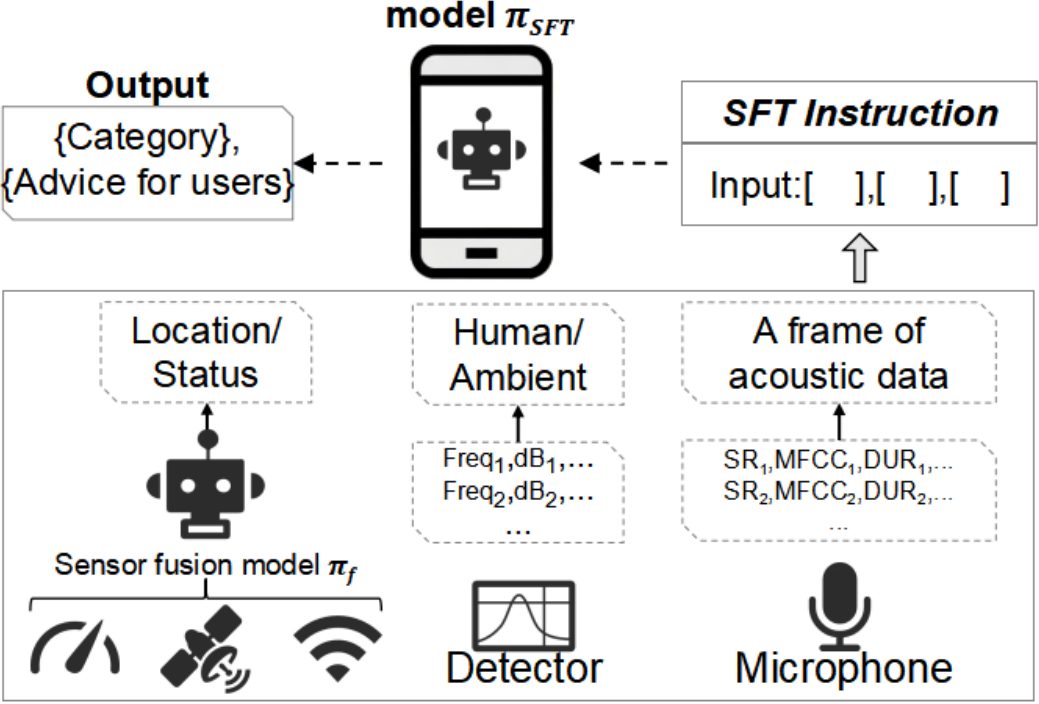}}
    \caption{Illustration of the stage-wise supervised fine-tuning.}
    \vspace{-5mm}
    \label{fig:sft}
\end{figure}

\textit{Stage-wise Supervised Fine-Tuning}.
SFT aims to provide each model with stable and interpretable capabilities before preference alignment.
We train the context inference model and scenario reasoning model separately using role-specific triplets $(instruction,input,output)$ (\figref{fig:sft}).
The context model $\pi_f$ takes sampled sensor cues as input and outputs structured physical states, such as \textit{indoor/outdoor} and \textit{stationary/mobile}.
The scenario reasoning model $\pi_{\mathrm{SFT}}$ takes inferred context labels, GMM-derived acoustic tags, and raw audio as input, and generates scenario-aware notifications and physical guidance.
The SFT objective is:
\begin{equation}
    \mathcal{L}_{\mathrm{SFT}} = -\sum_{t=1}^{T} \log P\left(y_t \mid x, y_{<t}\right),
\end{equation}
where $x$ denotes the input signals and instruction, $y_t$ represents the target token, and $T$ denotes the output length.
After SFT, \sysname obtains a grounded context model $\pi_f$ and an initial guidance model $\pi_{\mathrm{SFT}}$.
However, SFT mainly captures population-level reasoning patterns and cannot fully model how DHH users trade off urgency, interruption, and responsiveness across scenarios.

\textit{Scenario-Conditioned Preference Alignment}.
\label{sec:DPO}
DHH assistance requires scenario-dependent preferences:
the same sound may require different responses depending on whether the user is indoors or outdoors, stationary or mobile.
For example, a distant car horn outside a building may be safely ignored, while the same sound near a street crossing requires immediate attention.
Therefore, \sysname performs \textit{preference alignment} after scenario routing.
The structured prompt maps each instance into a scenario slot, including:
\textit{indoor/stationary},
\textit{indoor/mobile},
\textit{outdoor/stationary}, and
\textit{outdoor/mobile}.
This converts mixed-context preference learning into route-conditioned alignment.
Within each scenario slot, we model two key preference dimensions:
\textit{sound priority} and
\textit{accuracy--latency trade-off}.
Preference pairs are constructed from the DHH user study and scenario-specific guidance principles.
For each input $x$, the preferred response $y_w$ is concise, context-aware, and action-oriented.
The rejected response $y_l$ violates one or more principles, such as ignoring physical context, misjudging urgency, omitting recommended actions, or prioritizing inappropriate sounds.
Ambiguous cases are excluded from DPO training, while safety-critical events follow the deterministic override path described in \secref{sec:4.1.2}.

We adopt Direct Preference Optimization (DPO) instead of RLHF~\cite{bib:Rlhf}, as it avoids explicit reward modeling and rollout-based optimization.
Starting from $\pi_{\mathrm{SFT}}$, DPO optimizes the preference margin between $y_w$ and $y_l$:
\begin{equation}
\mathcal{L}_{\mathrm{DPO}}\left(\pi_{\theta} ; \pi_{\mathrm{ref}}\right)
=-\mathbb{E}\left[
\log \sigma\left(
\beta \log \frac{\pi_{\theta}(y_w \mid x)}{\pi_{\mathrm{SFT}}(y_w \mid x)}
-\beta \log \frac{\pi_{\theta}(y_l \mid x)}{\pi_{\mathrm{SFT}}(y_l \mid x)}
\right)\right]
\end{equation}
where $\pi_{\theta}$ is initialized from $\pi_{\mathrm{SFT}}$, $\pi_{\mathrm{ref}}=\pi_{\mathrm{SFT}}$, and $\beta$ controls deviation from the reference model.

\begin{figure}[t]
    \centering   
    \vspace{-5mm}
    \includegraphics[width=0.85\linewidth]{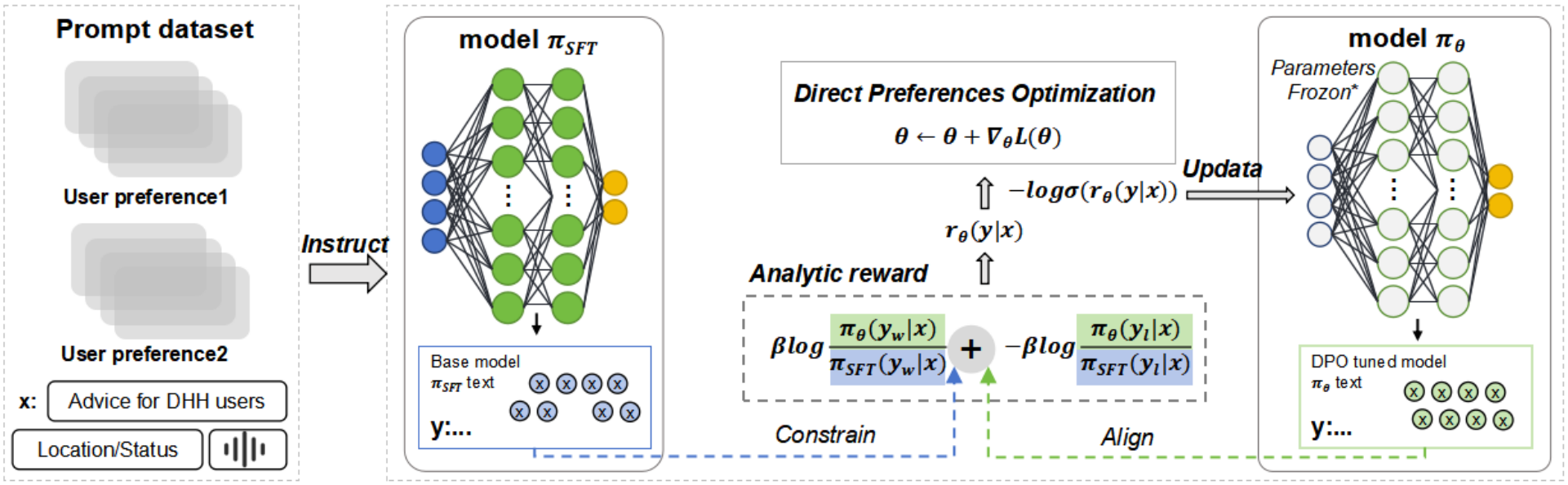}
    \caption{{Illustration of the scenario-conditioned preference alignment mechanism.}}
    \label{fig:dpo}
\end{figure}

\textit{Example Preference Cases}.
We illustrate scenario-conditioned DPO using non-safety-critical events (Table~\ref{tb:case}).
Safety-critical sounds such as sirens and alarms are excluded because they bypass LLM reasoning through deterministic override.
Across these cases, preferred responses are selected because they are concise, scenario-aware, and action-oriented, whereas rejected responses are context-insensitive, overly aggressive, or behaviorally misleading.
\begin{table}[h] 
\tiny
\centering 
\caption{Examples of scenario-conditioned preference pairs.} \label{tab:dpo_examples} \small \begin{tabular}{p{0.20\linewidth}p{0.36\linewidth}p{0.36\linewidth}} \toprule \textbf{Scenario} & \textbf{Preferred response $y_w$} & \textbf{Rejected response $y_l$} \\ \midrule Indoor, stationary; distant car horn & A distant car horn is detected outside. Since you are indoors and stationary, no immediate action is needed. & A car horn is detected. Stop what you are doing and immediately check for nearby traffic. \\ \midrule Outdoor, mobile; nearby car horn & A nearby car horn is detected. Stay alert, slow down, and check surrounding vehicles before moving. & A car horn is detected. It is common background noise and can be ignored. \\ \midrule Indoor, mobile; nearby speech & Nearby speech is detected while you are walking. Slow down and check whether someone is calling you. & Speech is detected indoors. It is likely irrelevant background conversation, so keep walking. \\ \midrule Outdoor, stationary; repeated honks & Repeated vehicle honks are detected nearby. Stay attentive and check traffic conditions before moving. & Repeated honks are normal urban noise. You can safely disregard them completely. 
\\ \bottomrule 
\end{tabular} 
\label{tb:case}
\end{table}

\subsection{Human-in-the-Loop On-Device Guidance Adaptation}
\label{sec:action}

Although \secref{sec:reason} aligns \sysname with population-level DHH preferences, assistive guidance remains user-specific. 
The same acoustic event may require different guidance across users, routines, and environments.
Existing personalization approaches often rely on confidence-triggered feedback or cloud-side retraining, which may miss confident-but-wrong decisions and introduce privacy and deployment overhead.
\sysname treats personalization as \textit{controlled decision correction} rather than full-model re-optimization.
It learns lightweight user-specific adjustments while preserving population-aligned reasoning through constrained on-device adaptation.

\subsubsection{Active Per-Instance Feedback}
Confidence alone cannot identify user-specific mismatches, as models may produce confident but inappropriate guidance.
Therefore, \sysname enables per-instance feedback after each response.
Users can correct both event interpretation and recommended actions, with each correction stored locally as a supervision pair for targeted adaptation.
This transforms personalization from global retraining into selective error correction.

\subsubsection{Efficient Adaptation with Output-Layer Adapter}
On-device personalization must adapt to individual preferences without destabilizing general reasoning.
Naive fine-tuning may overfit sparse feedback or overwrite previous corrections.
Thus, \sysname freezes the backbone models and updates only a lightweight output-layer adapter using LoRA-style low-rank updates.
This localizes adaptation at the decision boundary while preserving acoustic and scenario reasoning.
Formally, \sysname adjusts the output logits as:
\begin{equation}
\pi_{\text{final}}(y \mid x)
=
\pi_{\theta}(y \mid x)
+
\alpha \,\Delta_U(x),
\end{equation}
where $\pi_{\theta}$ denotes the frozen population model, $\Delta_U$ represents the user-specific adapter, and $\alpha$ controls adaptation strength.
The adapter is optimized with a stability-regularized objective:
\begin{equation}
\mathcal{L}_{\text{user}}
=
\mathcal{L}_{\text{DPO}}\!\left(\pi_{\text{final}}, \mathcal{D}_u\right)
+
\lambda \cdot \mathrm{KL}\!\left(
\pi_{\text{final}} \;\|\; \pi_{\theta}
\right)
+
\eta \cdot \mathcal{R}_{\text{stab}},
\end{equation}
where $\mathcal{D}_u$ is the local feedback set.
The KL constraint preserves population-aligned behavior, while $\mathcal{R}_{\text{stab}}$ prevents repeated rewriting of similar decisions.
Together, they make personalization a bounded correction rather than behavioral drift.
For efficiency, \sysname updates only FP16 adapter parameters.
A stabilization window after each update further reduces oscillation and self-overwriting during continuous adaptation.

\begin{figure}[t]
    \centering  
    \vspace{-5mm}
    \includegraphics[width=0.68\linewidth]{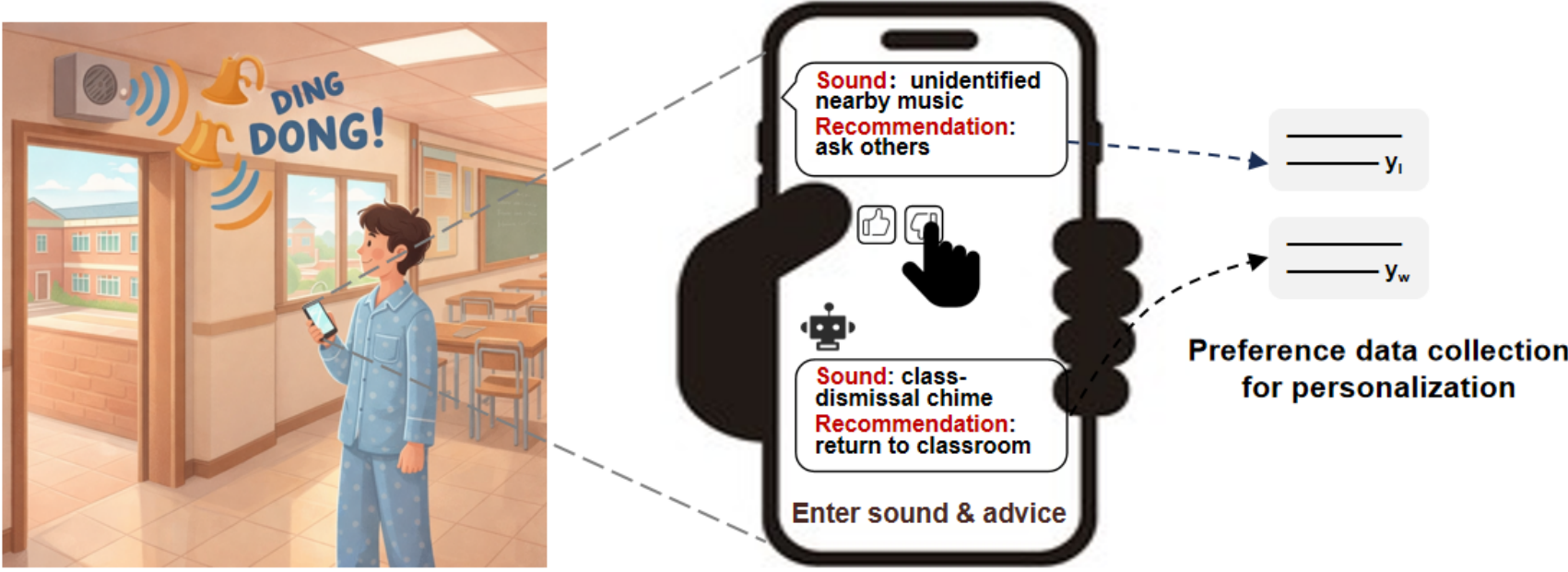}
    \caption{{Illustration of the customized configurations.}}
    \label{fig:diy}
    \vspace{-5mm}
\end{figure}

\subsubsection{Default and Customized Operation}
\sysname supports two modes. 
\sysname supports both default and customized operation.
In \textit{default} mode, pretrained models provide scenario-aware guidance without user intervention.
In \textit{customized} mode, user corrections trigger incremental local adapter updates.
As shown in \figref{fig:diy}, a high school student, \textit{Lucas}, initially receives generic guidance when broadcast music is interpreted as an undefined ambient sound.
After correcting it as a class-bell chime and specifying the desired action (\textit{``return to the classroom''}), \sysname learns this preference and provides more appropriate guidance for similar situations.

%% file: body/eval.tex
\section{EVALUATION}
This section presents the experimental settings and system performance of \sysname.

\subsection{Experiment Setups}
\label{sec:setup}

\fakeparagraph{Implementation}
\sysname is implemented as a two-stage pipeline. Offline training (SFT, DPO) and ablation studies are conducted on a server with an NVIDIA RTX 3090 GPU and 256GB RAM. 
The online system is deployed as an Android application on a Google Pixel 8 for continuous background sensing and multi-modal reasoning.
\wfm{Offline training and ablation studies use FP16, while on-device deployment uses 2-bit weight-only quantization for the reasoning modules.}
Table~\ref{tab:quantized_footprint} summarizes the quantized footprint of \sysname and the reasoning baselines.

\fakeparagraph{On-device deployment}
We deploy both \sysname modules on a Google Pixel 8 with 2-bit weight-only quantization. 
The 1.5B fusion model $\pi_f$ and the 1.8B audio-language model $\pi_\theta$ are co-located on device but executed sequentially in an event-driven pipeline: after a non-silent acoustic trigger, $\pi_f$ infers context labels, and $\pi_\theta$ generates guidance from these labels and the long-listened audio. 
Their quantized footprints are approximately 0.38\,GB and 0.45\,GB, respectively, making the combined stack feasible on Pixel 8-class hardware. For fairness, we apply the same 2-bit quantization to all LLM baselines, but report full Android deployment only for \sysname because deploying 7B--8B audio-LLM baselines requires additional model-specific kernel and memory optimizations.

\begin{table}[t]
\centering
\scriptsize
\caption{Quantized model footprint for reasoning modules.}
\label{tab:quantized_footprint}
\begin{tabular}{lccc}
\toprule
Model & Params & Quantization & Raw weight memory \\
\midrule
$\pi_f$ fusion model & 1.5B & 2-bit & $\sim$0.38GB \\
$\pi_\theta$ audio-language model & 1.8B & 2-bit & $\sim$0.45GB \\
\sysname reasoning stack & 3.3B & 2-bit & $\sim$0.83GB \\
Qwen-Audio2 & 7B & 2-bit & $\sim$1.75GB \\
Audio-Flamingo & 8B & 2-bit & $\sim$2.00GB \\
MiDashengLM & 8B & 2-bit & $\sim$2.00GB \\
\bottomrule
\end{tabular}
\end{table}

\fakeparagraph{Tasks and Models}
\lsc{We use a \textbf{self-collected physical context sensory dataset} from a Google Pixel 8 (1K+ clips over 7 days), including 18K accelerometer readings, 3.5K Wi-Fi AP counts, 2.8K SSID entries, and 4,200 satellite counts with SNR values, to supervise $\pi_f$.}
We also use AudioSet~\cite{bib:audioset} for acoustic training and large-scale evaluation. We evaluate \sysname on two tasks: coarse detection ($T_1$) and multi-modal context-aware guidance ($T_2$), measuring both task performance and on-device energy cost.

\fakeparagraph{Baselines}
Because \sysname covers both sensing and guidance generation, we compare it with representative baselines from two families.

\begin{itemize}
\item \textbf{Lightweight DNN-based sound detection methods} benchmark $T_1$ under different accuracy, latency, and energy trade-offs.
\begin{itemize}
    \item \textit{Silero}~\cite{bib:silero}: an edge-oriented neural voice activity detector with low parameter cost.
    \item \textit{YAMNet}~\cite{bib:yamnet}: a CNN-based audio event model from Google, pre-trained on large-scale audio data and used as a stronger but heavier detection baseline.
\end{itemize}

\item \textbf{LLM and non-LLM reasoning methods} benchmark $T_2$ for scenario-aware guidance.
\begin{itemize}
    \item \textit{SoundWatch}~\cite{bib:SoundWatch}: a non-LLM VGG-lite sound classification baseline. Since it outputs labels rather than guidance, we convert its predictions into fixed rule-based notifications for pairwise evaluation.
    \item \textit{Qwen-Audio2} (7B)~\cite{bib:qwen}: an audio-language model combining Whisper-large-v3 with Qwen-7B for speech, audio analysis, and mixed-scenario reasoning.
    \item \textit{Audio-Flamingo} (8B)~\cite{bib:flamingo}: an open-source audio-language framework supporting long-audio understanding across speech, music, and environmental sounds.
    \item \textit{MiDashengLM} (8B)~\cite{bib:midashenglm}: an audio-language model designed for low-resource acoustic scenarios and efficient zero-/few-shot audio-text generation.
\end{itemize}
\end{itemize}

Together, these baselines cover the key axes of \textit{efficiency \textit{vs.} generalization} and \textit{task-specific modeling \textit{vs.} general-purpose reasoning}, allowing us to evaluate how \sysname balances on-device efficiency with scenario-aware, preference-aligned assistance.

\fakeparagraph{Metrics}
We evaluate \sysname along three dimensions: 
(1) \textbf{Detection quality}, using Precision-Recall (PR) curves across SNR levels; 
(2) \textbf{System efficiency}, using on-device latency and energy consumption; and 
(3) \textbf{Guidance quality}, using pairwise preference win rates.
The win rate is obtained through blind pairwise comparisons, with Qwen3-Omni (30B)~\cite{bib:omni} as the primary automated judge for large-scale guidance evaluation.
\wfm{Human participants are involved separately in the 21-day field study and the 100-user quick test, where they assess perceived usefulness, understandability, and actionability rather than serving as judges for the large-scale pairwise benchmark.}

\begin{figure}[t]
    \centering
    \vspace{-5mm}
    \subfloat[Subset $D_1$]{\label{fig:T1}
        \includegraphics[width=0.23\linewidth]{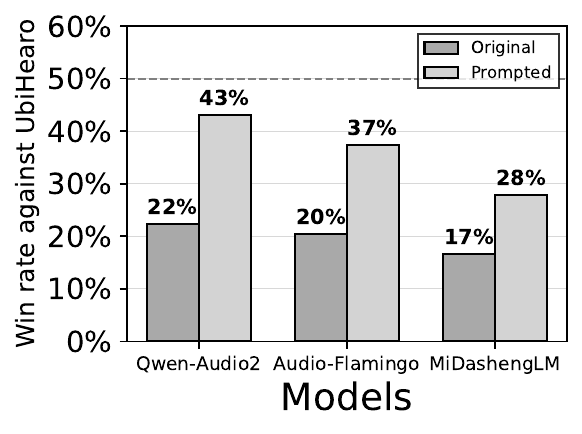}}
    \subfloat[Subset $D_2$]{\label{fig:T2}
        \includegraphics[width=0.23\linewidth]{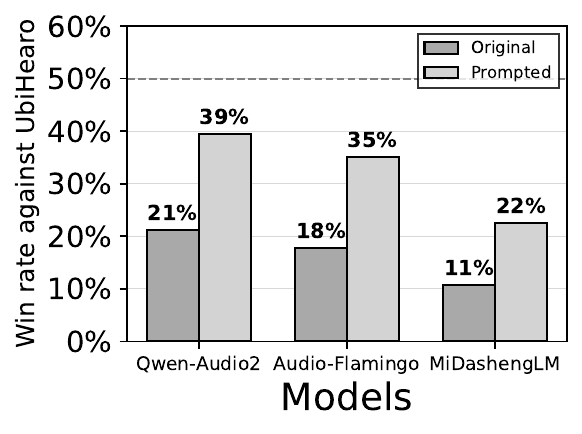}}
    \subfloat[Subset $D_3$]{\label{fig:T3}
        \includegraphics[width=0.23\linewidth]{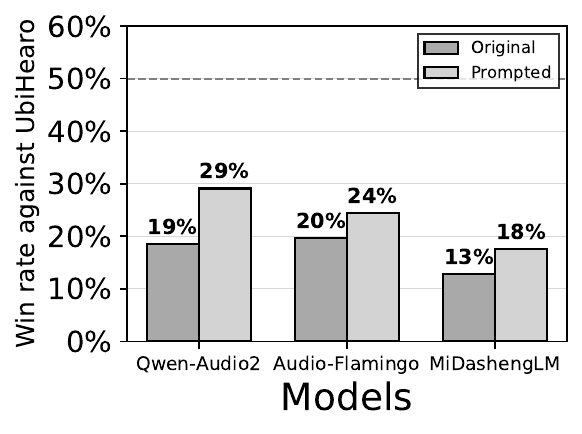}}
    \subfloat[Subset $D_4$ ]{\label{fig:T4}
        \includegraphics[width=0.23\linewidth]{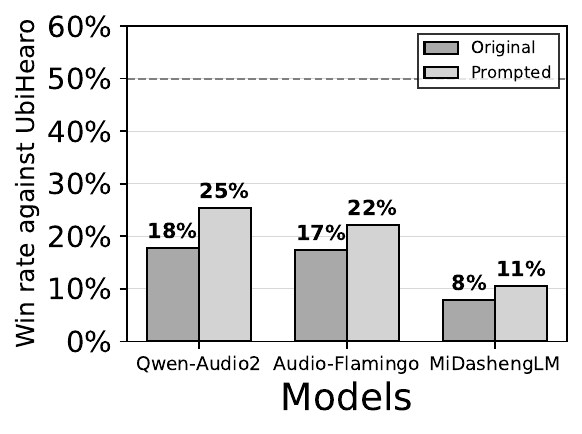}}
    \caption{\sysnameposs($\pi_\theta$) pairwise preference win rate on diverse scenarios.}
    \label{fig:winrate}
\end{figure}

\begin{figure}[t]
    \centering
    \vspace{-5mm}
    \subfloat[Subset $D_1$]{\label{fig:D1}
        \includegraphics[width=0.23\linewidth]{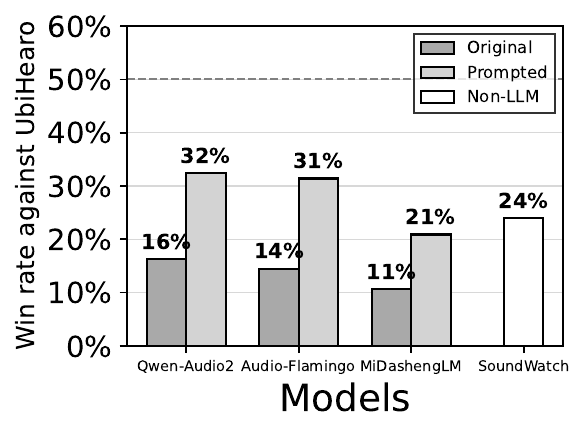}}
    \subfloat[Subset $D_2$]{\label{fig:D2}
        \includegraphics[width=0.23\linewidth]{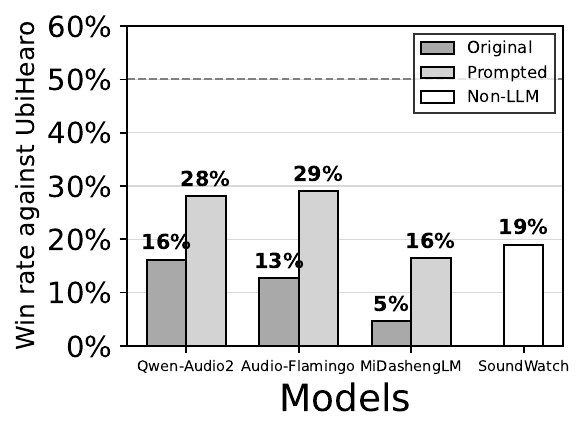}}
    \subfloat[Subset $D_3$]{\label{fig:D3}
        \includegraphics[width=0.23\linewidth]{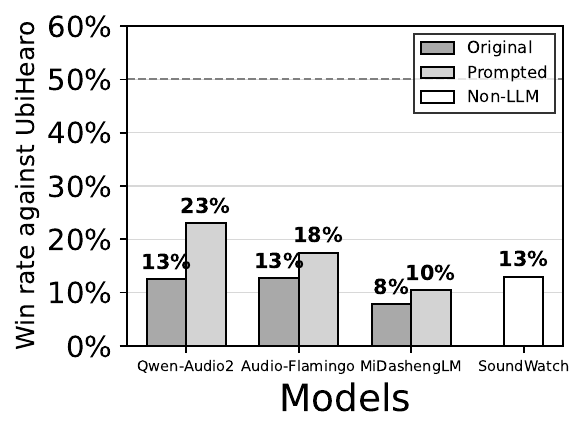}}
    \subfloat[ Subset $D_4$]{\label{fig:D4}
        \includegraphics[width=0.23\linewidth]{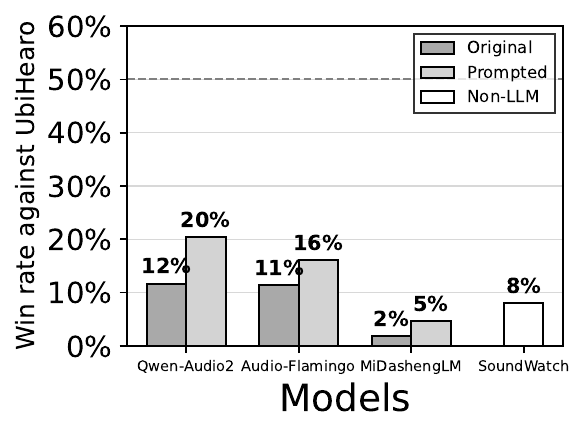}}
    \caption{\sysnameposs($\pi_f+\pi_\theta)$ pairwise preference win rate on diverse scenarios.}
    \vspace{-5mm}
    \label{fig:winrate1}
\end{figure}

\subsection{Performance of \sysnameposs Reasoning Module}
\label{sec:performance}
As described in \secref{sec:reason}, \sysname replaces monolithic audio-MLLMs with a \textit{two-stage} reasoning pipeline.
A compact modal-fusion model $\pi_f$ (1.5B) first infers structured user context from on-device sensors, and a lightweight audio-language model $\pi_\theta$ (1.8B) then performs scenario reasoning over long-listened audio.
The GMM acoustic tags and inferred context labels are used as structured prompts to $\pi_\theta$, with raw audio as the primary input.
We evaluate scenario-aware reasoning using \textit{pairwise preference win rate} (\secref{sec:setup}) on four AudioSet-derived subsets (2,000 clips): $D_1$ indoor/stationary isolated sounds, $D_2$ outdoor/mobile isolated sounds, $D_3$ indoor/stationary speech+ambient overlap, and $D_4$ outdoor/mobile speech+ambient overlap.
We compare (i) $\pi_\theta$ alone and (ii) the full pipeline ($\pi_f+\pi_\theta$) against Qwen-Audio2 (7B)~\cite{bib:qwen}, Audio-Flamingo (8B)~\cite{bib:flamingo}, and MiDashengLM (8B)~\cite{bib:midashenglm}, each in both original and prompted forms.

\begin{figure}[t]
    \centering  
    \includegraphics[width=0.7\linewidth]{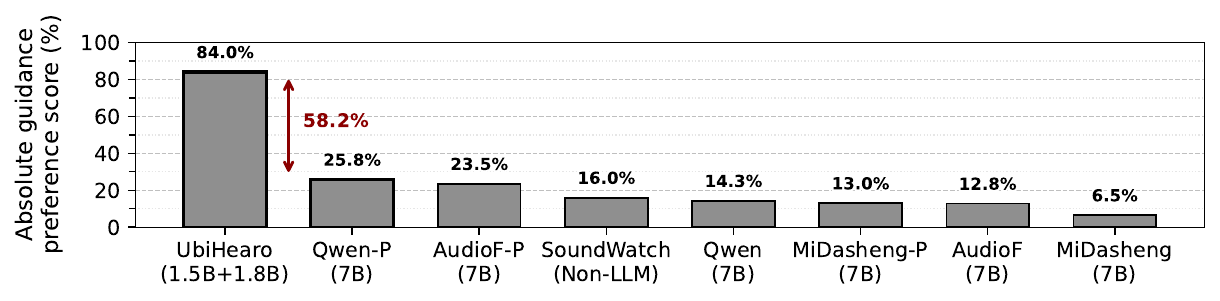}
    \caption{Absolute guidance preference score converted from the pairwise win rates in \figref{fig:winrate1}.
For each baseline $b$ on subset $D_i$, its score is its win rate against \sysname, \ie $S_{b,i}=W_{b,i}$.
For \sysname, we report the mean complementary win rate against all baselines,
$S_{\sysname,i}=\frac{1}{|\mathcal{B}|}\sum_{b\in\mathcal{B}}(100-W_{b,i})$.
The final score is averaged over $D_1$--$D_4$.
``P'' denotes the prompted version.
Higher is better.
}
    \label{fig:absolute_guidance_score}
    \vspace{-2mm}
\end{figure}

\subsubsection{Performance of Scenario Reasoning Model $\pi_\theta$}
\label{sec:expre1}
We first evaluate $\pi_\theta$ under identical inputs and prompts (\textit{GMM outputs + context labels}).
\wfm{We conduct blind pairwise preference evaluation using Qwen3-Omni as the automated judge, where the judge selects the more useful and actionable guidance for each clip under randomized output order.}
As shown in \figref{fig:winrate}, \sysname consistently exceeds 50\% win rate against all baselines.
On $D_1$ and $D_2$, it wins 57\% and 61\% against the strongest baseline, prompted Qwen-Audio2.
The gain is larger in overlapped and dynamic settings: on $D_4$, \sysname reaches up to 92\% win rate, while MiDashengLM drops to 8\%.
These results show that SFT+DPO enables $\pi_\theta$ to produce more scenario-aware and preference-aligned guidance under challenging acoustic conditions.

\subsubsection{Performance of \sysnameposs Two-stage Pipeline ($\pi_f$+$\pi_\theta$)}
\label{sec:expre2}
We then evaluate the full pipeline on the same four subsets.
All LLM-based methods use the same long-listened audio and prompts containing \textit{GMM outputs + raw on-device sensor readings}.
Unlike the baselines, \sysname first converts raw sensors into structured context labels through $\pi_f$, testing whether explicit context grounding improves guidance beyond model scale.
\lsc{We also include a SoundWatch-style non-LLM baseline~\cite{bib:SoundWatch}.
Since SoundWatch outputs labels rather than guidance, we convert labels into fixed notifications: \textit{``Detected [sound label]. Please check the source if needed.''}
We further add SoundWatch+Rule, which maps selected safety-related labels to simple actions, such as alarm/siren/horn $\rightarrow$ check surroundings, door knock/doorbell $\rightarrow$ check the door, water running $\rightarrow$ check the source, and otherwise $\rightarrow$ be aware.}
\lsc{As shown in \figref{fig:winrate1}, \sysname outperforms all baselines on every subset.
On $D_1$, it wins 68\% against the strongest LLM baseline, which reaches 32\%.
On $D_2$, \sysname achieves 71\%--95\% win rates under more dynamic conditions.
Against the SoundWatch-style baseline, \sysname wins 76\%, 81\%, 87\%, and 92\% on $D_1$--$D_4$, respectively, with larger gains in overlapped and outdoor-mobile scenarios where label-only or rule-based notifications struggle to infer urgency.
For clearer interpretation, we convert the win rates in \figref{fig:winrate1} into an absolute guidance preference score (\figref{fig:absolute_guidance_score}).
\sysname achieves 84.0\%, substantially higher than the strongest baseline, prompted Qwen-Audio2, at 25.8\%.}

\fakeparagraph{Summary}
Overall, \sysname improves DHH-oriented scenario understanding through structured context grounding and preference-aligned sensing--reasoning--action, enabling a compact two-stage model to outperform larger audio-MLLMs and non-LLM baselines.

\subsection{Deep Dive into \sysname's Two-stage Reasoning Scheme}
\label{sec:scheme}
As described in \secref{sec:pl}, \sysname uses stage-wise SFT to separate context understanding from guidance generation.
In Stage~1, the context model $\pi_f$ maps on-device sensor signals to structured context labels.
In Stage~2, the guidance model $\pi_{\text{SFT}}$ takes a structured prompt, composed of coarse acoustic tags and context labels, together with long-listened audio clips to generate scenario-aware guidance.
By contrast, Scheme~1 directly feeds \textit{sensor signals + short-listened audio} into a single-stage model, while Scheme~2 uses \textit{acoustic tags + sensor signals + long-listened audio} without explicit context structuring.

\subsubsection{Training Efficiency.}
We compare \sysname with Scheme~1 and Scheme~2 by fine-tuning the same base model (Qwen2-Audio-7B-Instruct) under identical SFT settings.
As shown in \figref{fig:loss}, both $\pi_f$ and $\pi_{\text{SFT}}$ converge within \textit{3 epochs}, faster than Scheme~1/2, and reach lower losses (0.18 and 0.36) than Scheme~1 (0.59) and Scheme~2 (1.11).
This shows that decomposing reasoning into context structuring and audio-grounded guidance reduces learning difficulty and improves training efficiency.

\subsubsection{Downstream Reasoning Quality.}
\begin{figure}[t]
    \centering
    \vspace{-5mm}
    \subfloat[The loss trajectories of SFT]{\label{fig:loss}
        \includegraphics[width=0.23\linewidth]{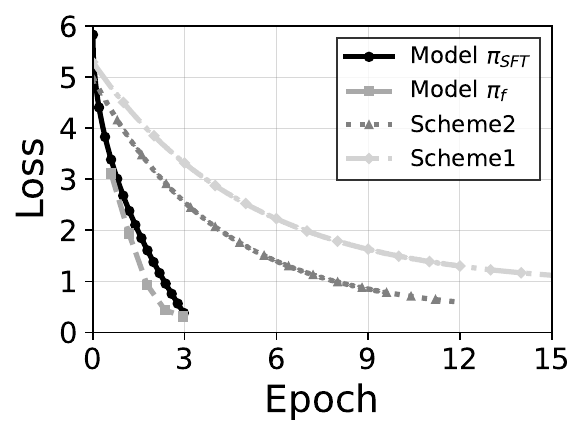}}
    \subfloat[Diverse schemes]{\label{fig:scheme}
        \includegraphics[width=0.23\linewidth]{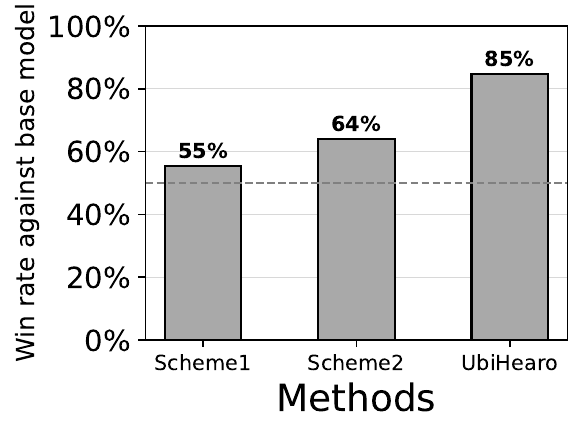}}
    \caption{Performance of \sysnameposs two-stage reasoning scheme.}
    \vspace{-5mm}
    \label{fig:schemes}
\end{figure}

We further evaluate guidance quality using pairwise preference win rate across Scheme~1, Scheme~2, \sysname, and the base model (\figref{fig:scheme}).
Despite using fewer parameters (1.5B$+$1.8B \textit{vs.} 7B), \sysname achieves an 84.7\% win rate over the base model, showing that two-stage reasoning with automatically generated structured prompts improves preference-aligned guidance beyond single-stage prompting.

\fakeparagraph{Summary}
Overall, \sysnameposs two-stage reasoning scheme improves both training convergence and scenario-aware guidance quality.

\subsection{Performance of \sysnameposs Acoustic Event Detector}
We evaluate \sysname's acoustic event detector on a \textit{Google Pixel~8} against two mobile baselines, YAMNet~\cite{bib:yamnet} and Silero~\cite{bib:silero}, focusing on noise robustness and always-on efficiency.
We test three SNR levels (15/20/25~dB) using 14,800 audio frames from 50 AudioSet clips covering \textit{Human Speech} and \textit{Ambient Sound}.

\subsubsection{Detection Accuracy under Different SNRs}
\figref{fig:vad1}-\figref{fig:vad3} report precision-recall curves and average precision (AP).
\sysname achieves AP scores of 0.922/0.954/0.991 across 15/20/25~dB, matching or outperforming both baselines at 20--25~dB and remaining slightly below YAMNet only at the near-threshold 15~dB condition.
Its runtime memory footprint is only 158~KB, \(\sim\)369\(\times\) smaller than YAMNet (57~MB), supporting continuous sensing under tight mobile resource budgets.

\subsubsection{Energy-Accuracy Trade-off}
\begin{figure}[t]
    \centering
    \vspace{-5mm}
    \subfloat[PR curve under 15dB]{\label{fig:vad1}
        \includegraphics[width=0.23\linewidth]{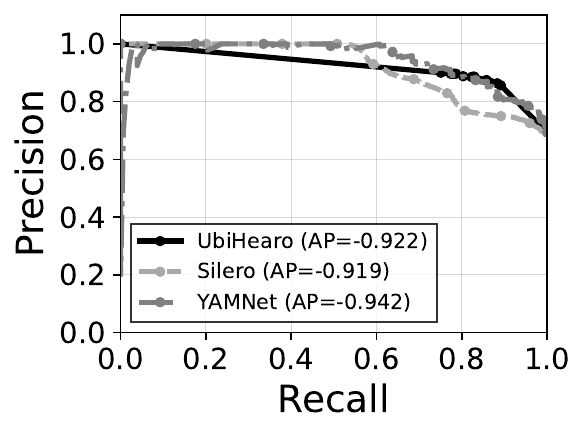}}
    \subfloat[PR curve under 20dB]{\label{fig:vad2}
        \includegraphics[width=0.23\linewidth]{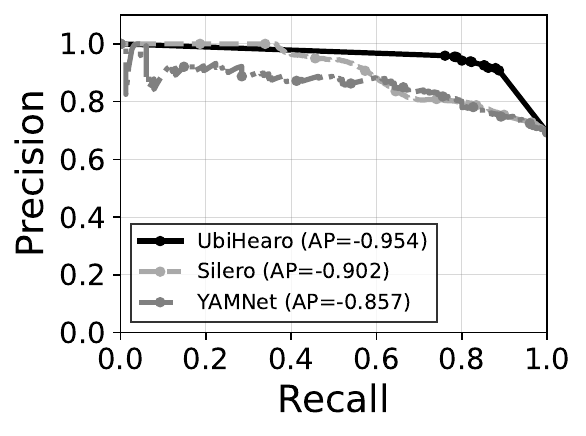}}
    \subfloat[PR curve under 25dB]{\label{fig:vad3}
        \includegraphics[width=0.23\linewidth]{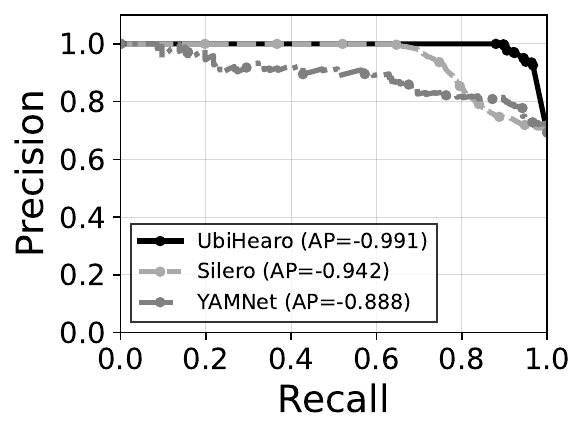}}
    \subfloat[Consumption vs. Duration]{\label{fig:vade}
        \includegraphics[width=0.23\linewidth]{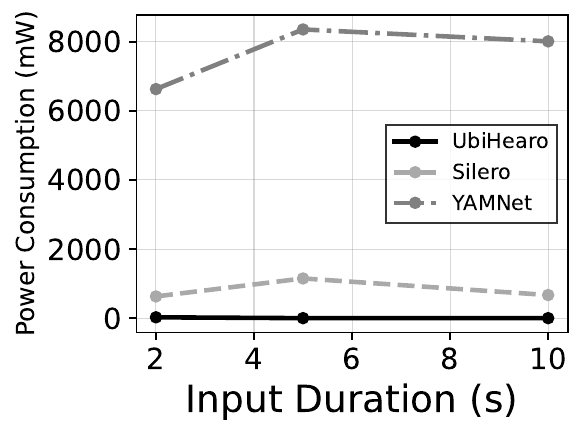}}
    \caption{Performance of acoustic event detector.}
    \vspace{-5mm}
    \label{fig:vad}
\end{figure}

We further measure on-device power at 20~dB SNR and compare \sysname with YAMNet and Silero.
As shown in \figref{fig:vade}, \sysname provides the best accuracy-energy trade-off: it consumes only 20~mW, \(\sim\)391\(\times\) lower than YAMNet (7{,}836~mW), while maintaining competitive detection precision (AP $\approx$ 0.956 averaged across SNRs).
\wfm{Energy is measured on a Google Pixel~8 using \textit{Android Battery Historian} and \textit{dumpsys batterystats} under the same always-on sensing condition.
Before each run, we reset battery statistics and closed unrelated background applications.}
These results show that \sysname enables accurate and energy-efficient coarse acoustic detection for always-on mobile sensing.

\subsection{End-to-end System Performance}
We evaluate \sysname's end-to-end overhead on Google Pixel~8 (4575~mAh, 8~GB RAM) and Pixel~9 (4700~mAh, 12~GB RAM) to examine its feasibility for \emph{always-on} mobile sensing.

\subsubsection{Power Consumption and Battery Life}
\label{sec:ove}
\begin{figure}[t]
    \centering
    \vspace{-5mm}
    \subfloat[Once/duty-cycling at pixel~8]{\label{fig:c81}
        \includegraphics[width=0.23\linewidth]{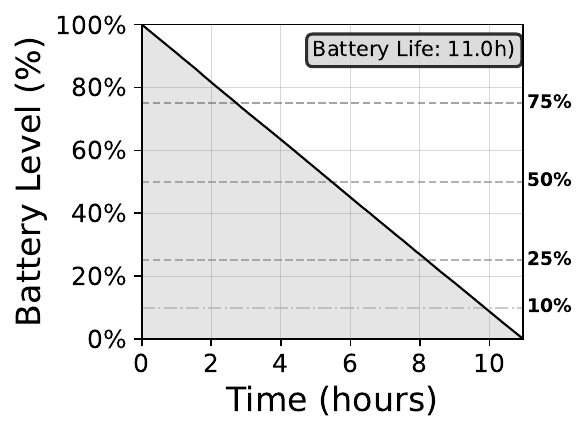}}
    \subfloat[Twice/duty-cycling at pixel~8]{\label{fig:c82}
        \includegraphics[width=0.23\linewidth]{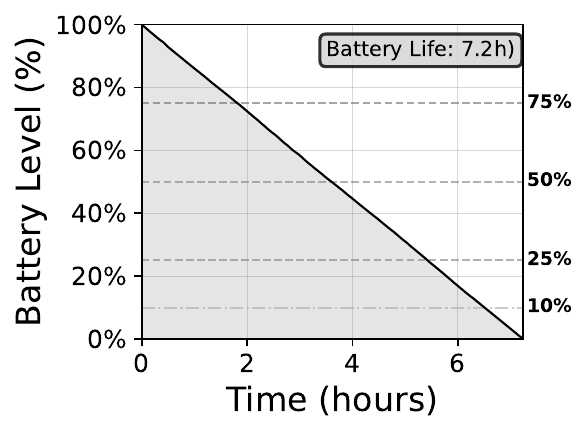}}
    \subfloat[Once/duty-cycling at pixel~9]{\label{fig:c91}
        \includegraphics[width=0.23\linewidth]{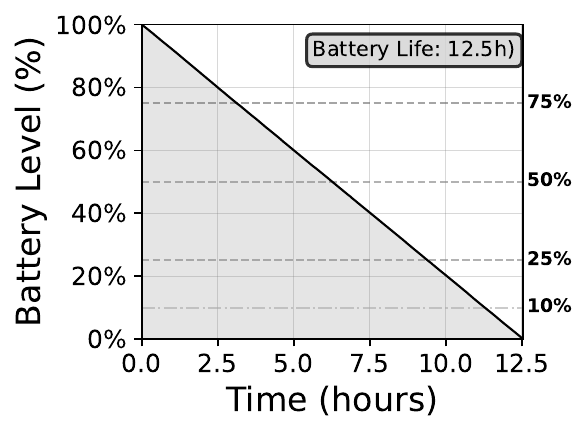}}
    \subfloat[Twice/duty-cycling at pixel~9]{\label{fig:c92}
        \includegraphics[width=0.23\linewidth]{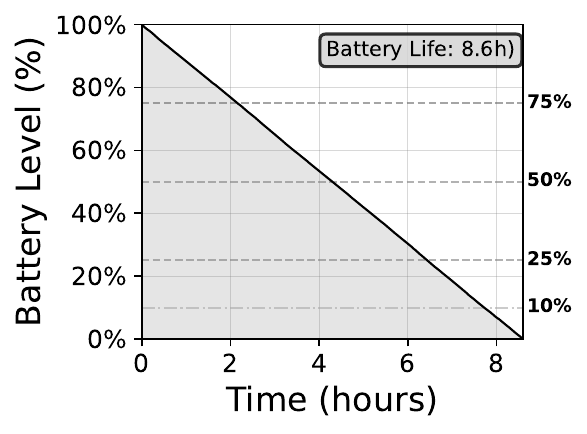}}
    \caption{System overhead of \sysname at Google pixel~8 and pixel~9 smartphones.}
    \vspace{-5mm}
    \label{fig:consump}
\end{figure}

We measure end-to-end power consumption by repeatedly triggering \sysname with locally stored AudioSet clips, avoiding network effects.
Specifically, we use ten 3-s \textit{Human Speech} clips and ten 3-s \textit{Ambient Sound} clips, keep the Pixel~8 screen off unless guidance is displayed, and randomly pair each trigger with one of four scenario-specific sensor instances for context inference by $\pi_f$.
We test two conservative trigger frequencies, once and twice per duty cycle, both more frequent than typical daily acoustic events.
As shown in \figref{fig:c81} and \figref{fig:c91}, \sysname achieves 11.0~h battery life on Pixel~8 and 12.5~h on Pixel~9 under once-per-duty-cycle triggering.
When triggered twice per duty cycle, battery life decreases to 7.2~h and 8.6~h, respectively (\figref{fig:c82}).
These results indicate that \sysname can support practical always-on use.

\subsubsection{System Latency}
\label{sec:lat}
\begin{figure}[t]
    \centering
    \vspace{-5mm}
    \subfloat[Latency at pixel~8]{\label{fig:lat1}
        \includegraphics[width=0.23\linewidth]{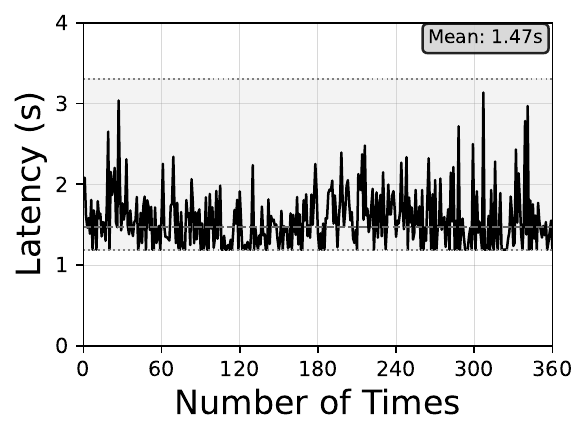}}
    \subfloat[Latency at pixel~9]{\label{fig:lat2}
        \includegraphics[width=0.23\linewidth]{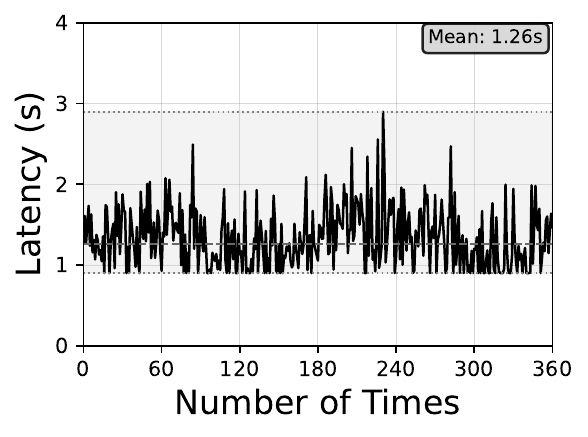}}
    \caption{Latency of \sysname at Google pixel~8 and pixel~9.}
    \label{fig:lat}
\end{figure}

We report end-to-end processing latency under the once-per-duty-cycle setting, measured between the fourth and fifth hour of the battery test to capture steady-state performance.
As shown in \figref{fig:lat1} and \figref{fig:lat2}, \sysname achieves near-real-time average latency: 1.47~s on Pixel~8 and 1.26~s on Pixel~9.
Existing audio-MLLM baselines, such as Qwen-Audio2, Audio-Flamingo, and MiDashengLM, are too large for fully on-device always-on deployment and therefore cannot provide comparable latency under the same setting.
Latency also varies by scenario: it is lowest for \textit{outdoor/mobile} cases, reaching 1.2~s on Pixel~8 and 0.9~s on Pixel~9, and highest for \textit{indoor/stationary} cases, reaching 3.3~s and 2.20~s, respectively.
This is because \sysname generates more concise guidance in outdoor/mobile scenarios and more detailed responses in indoor/stationary scenarios, as described in \secref{sec:DPO}.
\wfm{While \sysname achieves near real-time latency on the tested Pixel devices, its latency may still vary with device capability, runtime workload, and response length. Future work will further optimize scheduling and output-length control towards real-time processing across broader deployment settings.}

\fakeparagraph{Summary}
\sysname achieves energy-efficient always-on operation and near-real-time on-device performance on commodity smartphones, enabling timely notification in everyday use.

\subsection{Micro-Benchmark}
We conduct micro-benchmarks to quantify key design trade-offs in \sysname.
We concatenate 10-s AudioSet clips into long streams and randomly insert 2--10~s silent gaps to emulate sparse always-on listening.
The streams include speech, alarms, and short transient events, and we report \textit{Event Detection Recall}, defined as the fraction of ground-truth events successfully detected.

\subsubsection{Frame Length}
\label{sec:fl}
We evaluate four frame lengths (5--40~ms) to balance temporal granularity and stability (\figref{fig:b1}).
Short frames (5/10~ms) suffer from high energy variance and lower recall (75.1\%/86.3\%), while longer frames improve robustness.
Although 40~ms slightly improves recall over 20~ms (97.2\% \textit{vs.} 95.6\%), it doubles detection latency.
We therefore adopt 20~ms as the accuracy-latency trade-off.

\subsubsection{Duty-cycled Schedule}
\label{sec:dcs}
We evaluate 2--5 sampling phases per 10~s interval (\figref{fig:b2}).
Sparse schedules with 2/3 phases miss transient events and reduce recall (80.5\%/86.2\%).
Four phases increase recall to 95.7\%, while five phases bring only marginal gain (97.1\%) with higher overhead.
Thus, we use four phases by default.

\subsubsection{Listening Duration}
\label{sec:ld}
\begin{figure}[t]
    \centering
    \vspace{-5mm}
    \subfloat[Frame length]{\label{fig:b1}
        \includegraphics[width=0.23\linewidth]{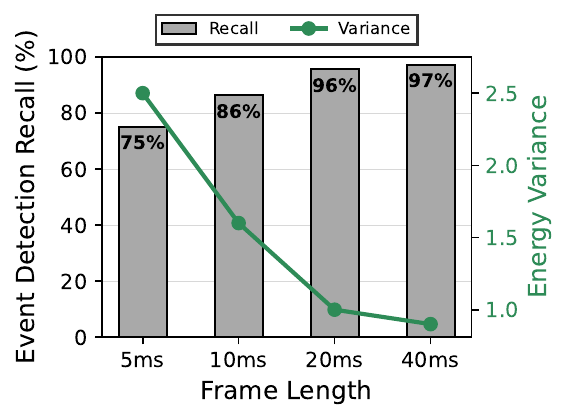}}
    \subfloat[Duty-cycling configurations]{\label{fig:b2}
        \includegraphics[width=0.23\linewidth]{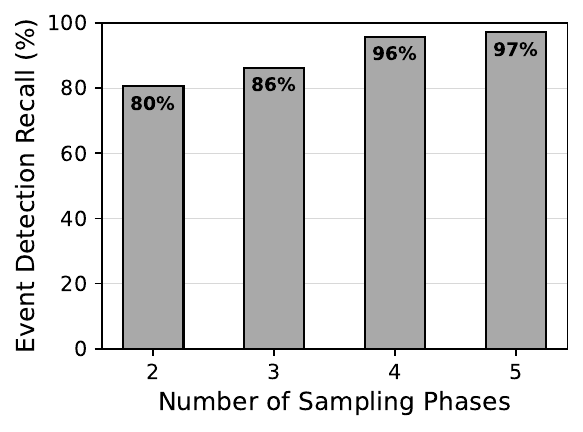}}
    \subfloat[Listening duration]{\label{fig:b3}
        \includegraphics[width=0.23\linewidth]{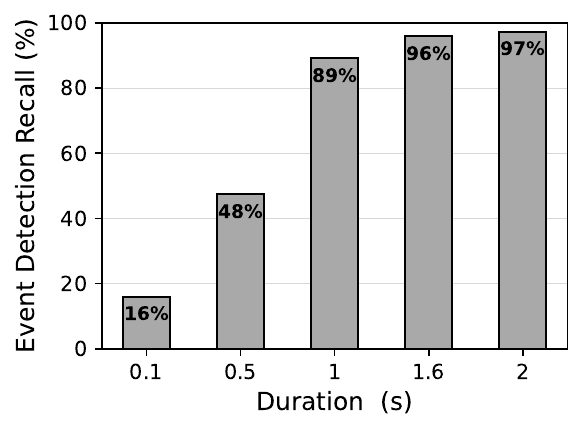}}
    \caption{Micro-benchmark of \sysnameposs configurations.}
    \vspace{-5mm}
    \label{fig:ben}
\end{figure}

We vary the listening duration from 0.1--2~s per trigger (\figref{fig:b3}).
A 1~s window already provides strong recall and is used in power-saving mode (battery $\leq$ 50\%).
A 1.6~s window improves recall to 95.9\% and is used in standard mode.
Thus, \sysname adopts an adaptive policy: 1~s for low-power operation and 1.6~s for higher-fidelity sensing.

\fakeparagraph{Summary}
These micro-benchmarks show that 20~ms frames, four duty-cycled phases, and adaptive listening duration provide a practical balance among recall, latency, and energy cost.

\begin{figure}[t]
    \centering
    \vspace{-5mm}
    \subfloat[Guidance satisfaction]{\label{fig:u1}
        \includegraphics[width=0.23\linewidth]{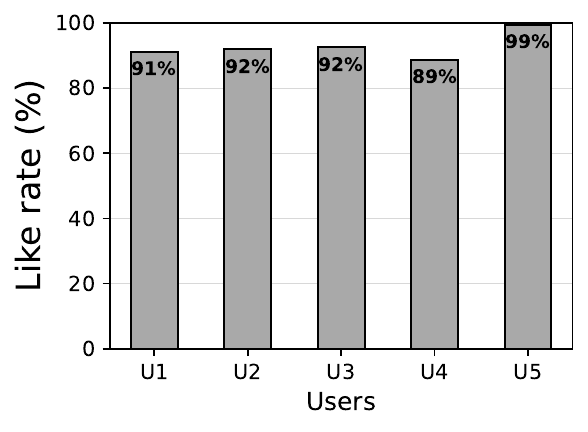}}
    \subfloat[Latency satisfaction]{\label{fig:u2}
        \includegraphics[width=0.23\linewidth]{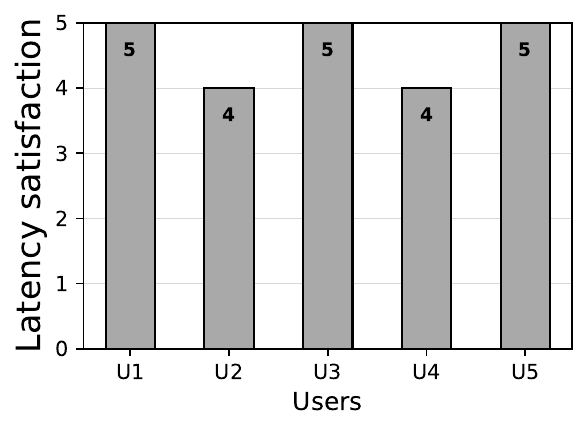}}
    \subfloat[Adaptation effect]{\label{fig:u3}
        \includegraphics[width=0.23\linewidth]{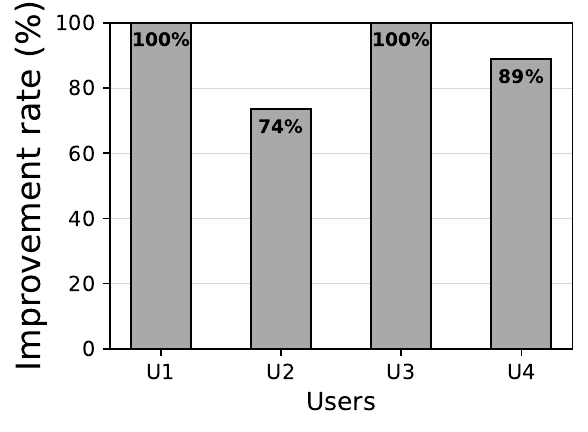}}     
    \vspace{-1mm}
    \subfloat[Examples of the questionnaire and user study process]{\label{fig:aid}
        \includegraphics[width=0.6\linewidth]{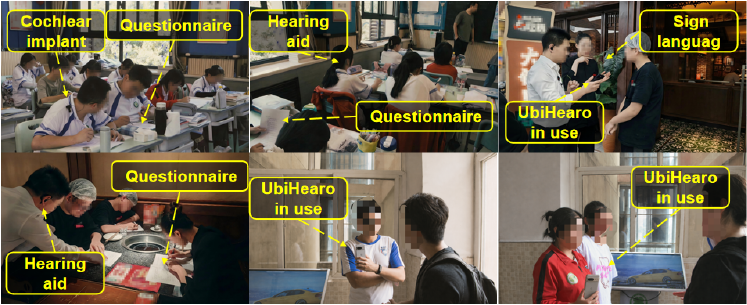}}
    \vspace{-2mm}
    \caption{User study.}
    \vspace{-3mm}
    \label{fig:user}
\end{figure}

\subsection{User Study}
We conducted a two-tier user study to evaluate whether \sysname provides useful guidance in daily life, not only accurate predictions in controlled tests.
\lsc{We employed five DHH participants for a 21-day longitudinal field study and 100 additional participants for a breadth-side quick test.}

\lsc{
\fakeparagraph{21-Day Daily Usage}
We employed five DHH participants: \textit{U1} (age 19, congenitally deaf, student), \textit{U2} (age 32, hard of hearing, employed), \textit{U3} (age 27, hard of hearing, homemaker), \textit{U4} (age 45, congenitally deaf, self-employed), and \textit{U5} (age 55, hard of hearing, unemployed).
We asked them to use \sysname in daily life for 21 days.
During the study, \sysname was triggered \textit{8,834}, \textit{9,443}, \textit{5,278}, \textit{6,811}, and \textit{3,514} times for \textit{U1}-\textit{U5}, respectively.
To reduce burden, we asked participants to tap the on-screen ``\textit{dislike}'' button (\figref{fig:diy}) only when the output did not match the current situation.
As shown in \figref{fig:u1}, \sysname achieved high user-rated appropriateness over 21 days: 91.2\% (8058/8834), 91.9\% (8670/9443), 92.5\% (4884/5278), 88.7\% (6043/6811), and 99.4\% (3492/3514) for \textit{U1}--\textit{U5}.
We attribute the variation across users to differences in daily environments.
As shown in \figref{fig:u2}, participants rated \sysnameposs latency as 5, 4, 5, 4, and 5, with an average score of 4.6.}

\fakeparagraph{Human-in-the-loop On-device Adaptation}
We asked four participants (\textit{U1}--\textit{U4}) to enable the \textit{customized} mode and adapt \sysname to personal sound cues that generic datasets rarely cover, including a campus chime for class transitions (\textit{U1}), TV audio at home (\textit{U2}), microwave sounds (\textit{U3}), and cash-register clicks at checkout (\textit{U4}).
As shown in \figref{fig:u3}, adaptation improved recognition of these personalized events: \textit{U1} 0/22$\rightarrow$23/23, \textit{U2} 0/30$\rightarrow$25/34, \textit{U3} 0/3$\rightarrow$3/3, and \textit{U4} 0/64$\rightarrow$32/36.
These results show that on-device human-in-the-loop adaptation helps \sysname learn user-specific sound meanings in real use.

\fakeparagraph{100-User Test}
We employed 100 additional participants from campus and local communities for a lightweight quick test.
We asked each participant to review four representative guidance clips, one for each scenario slot: indoor/stationary, indoor/mobile, outdoor/stationary, and outdoor/mobile.
This yielded 400 clip-level judgments.
For each clip, we asked participants to rate whether the scenario understanding was correct and whether the guidance was actionable.
Overall, \sysname achieved 89.5\% scenario-correct judgments (358/400) and 87.3\% actionable judgments (349/400), showing that its guidance remained understandable and actionable beyond the five long-term participants.
\wfm{The 100-user quick test complements the 21-day field study by providing a broader controlled review of representative guidance clips.
While the field study captures repeated real-world use, the quick test evaluates whether the guidance is understandable and actionable for a broader participant pool.
}

\begin{figure}[t]
    \centering
    \vspace{-5mm}
    \subfloat[Railway station]{\label{fig:rail}
        \includegraphics[width=0.2\linewidth]{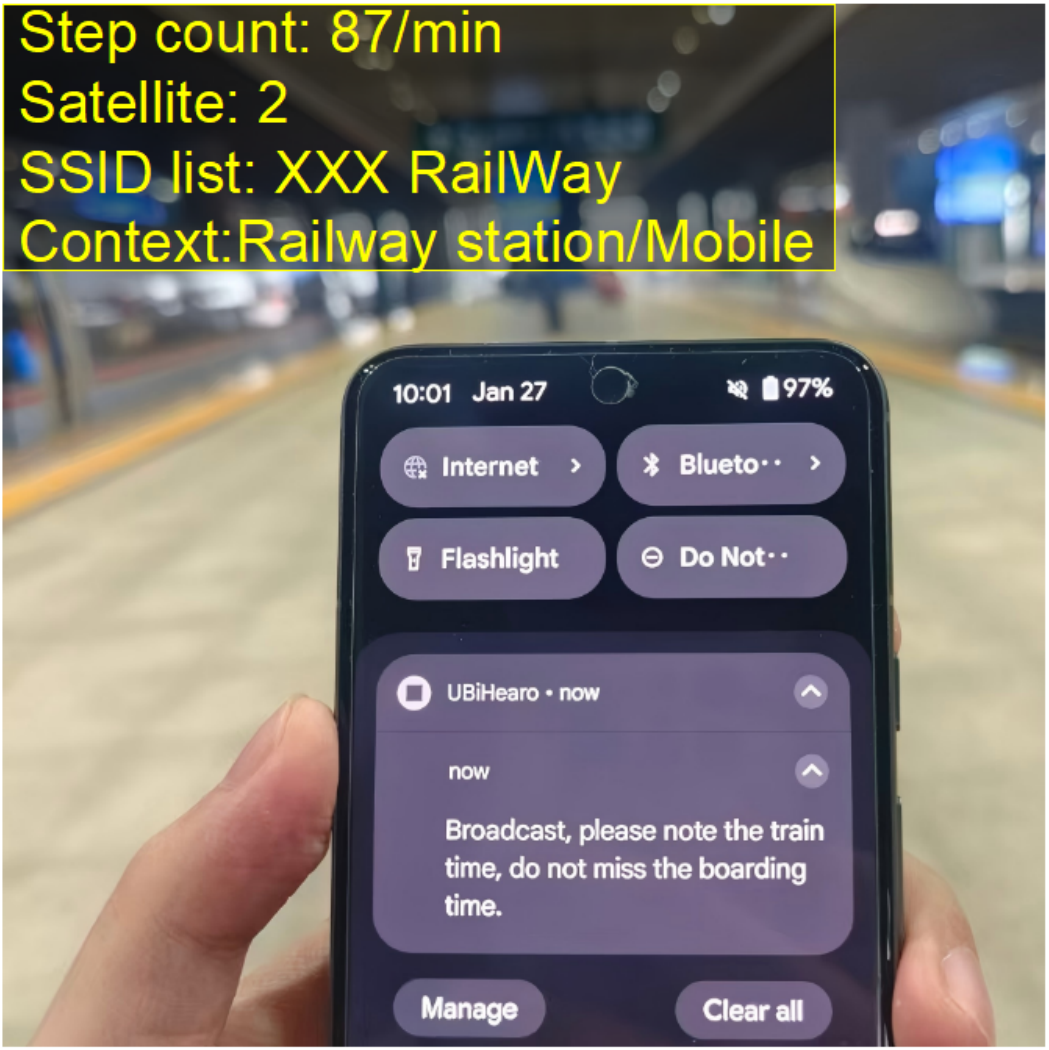}}
    \subfloat[Subway]{\label{fig:alert}
        \includegraphics[width=0.2\linewidth]{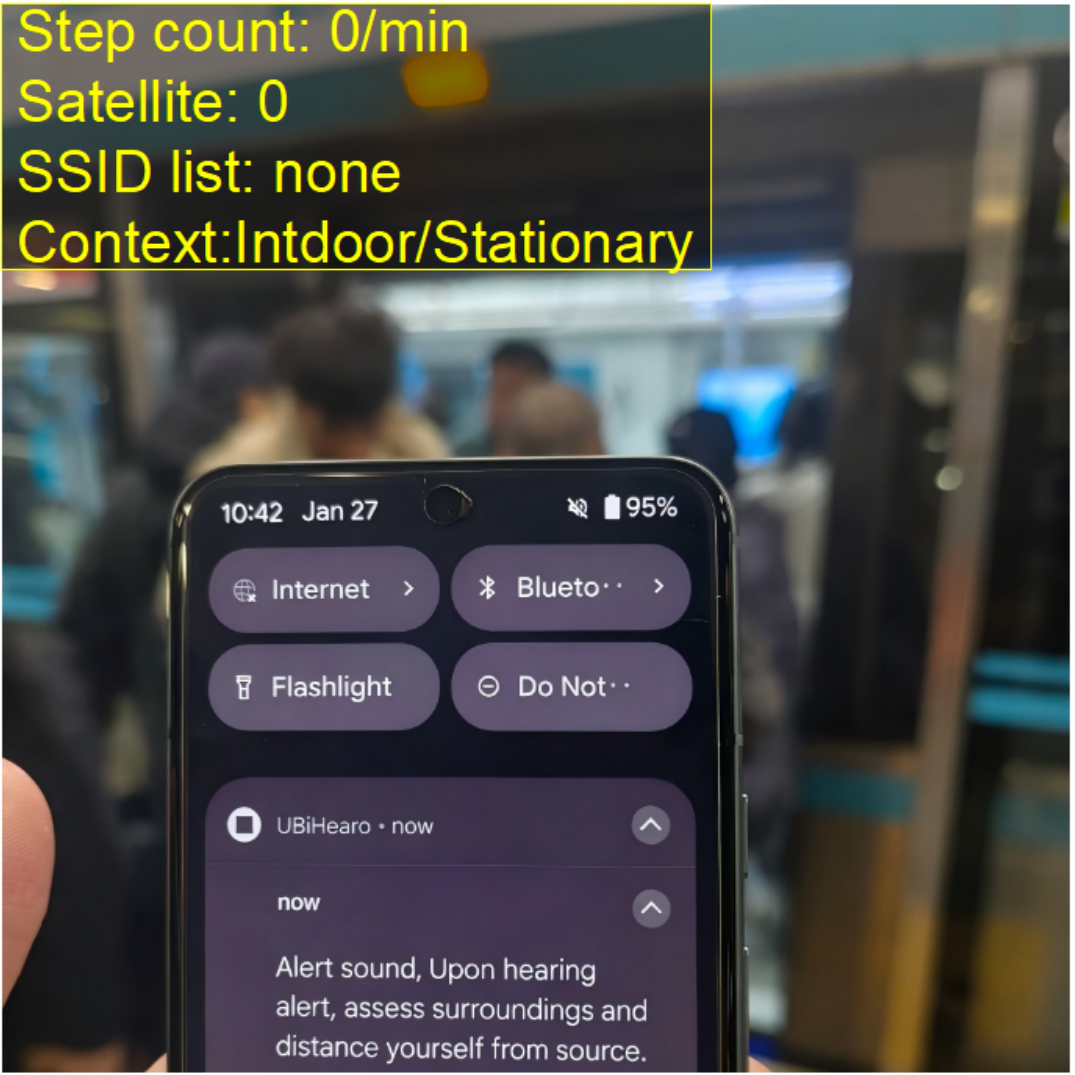}}
    \subfloat[Airport, walking]{\label{fig:air1}
        \includegraphics[width=0.2\linewidth]{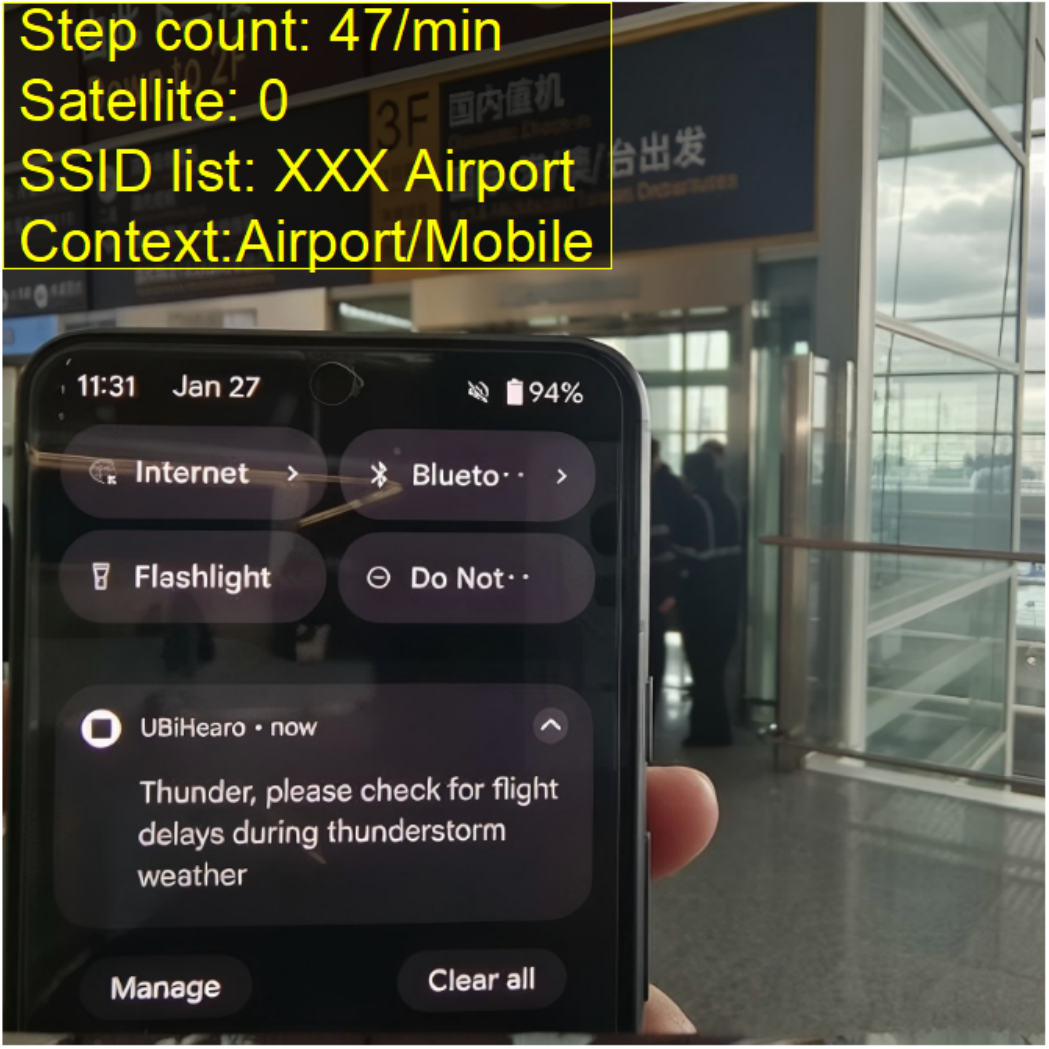}}
    \subfloat[Airport, stationary]{\label{fig:air2}
        \includegraphics[width=0.2\linewidth]{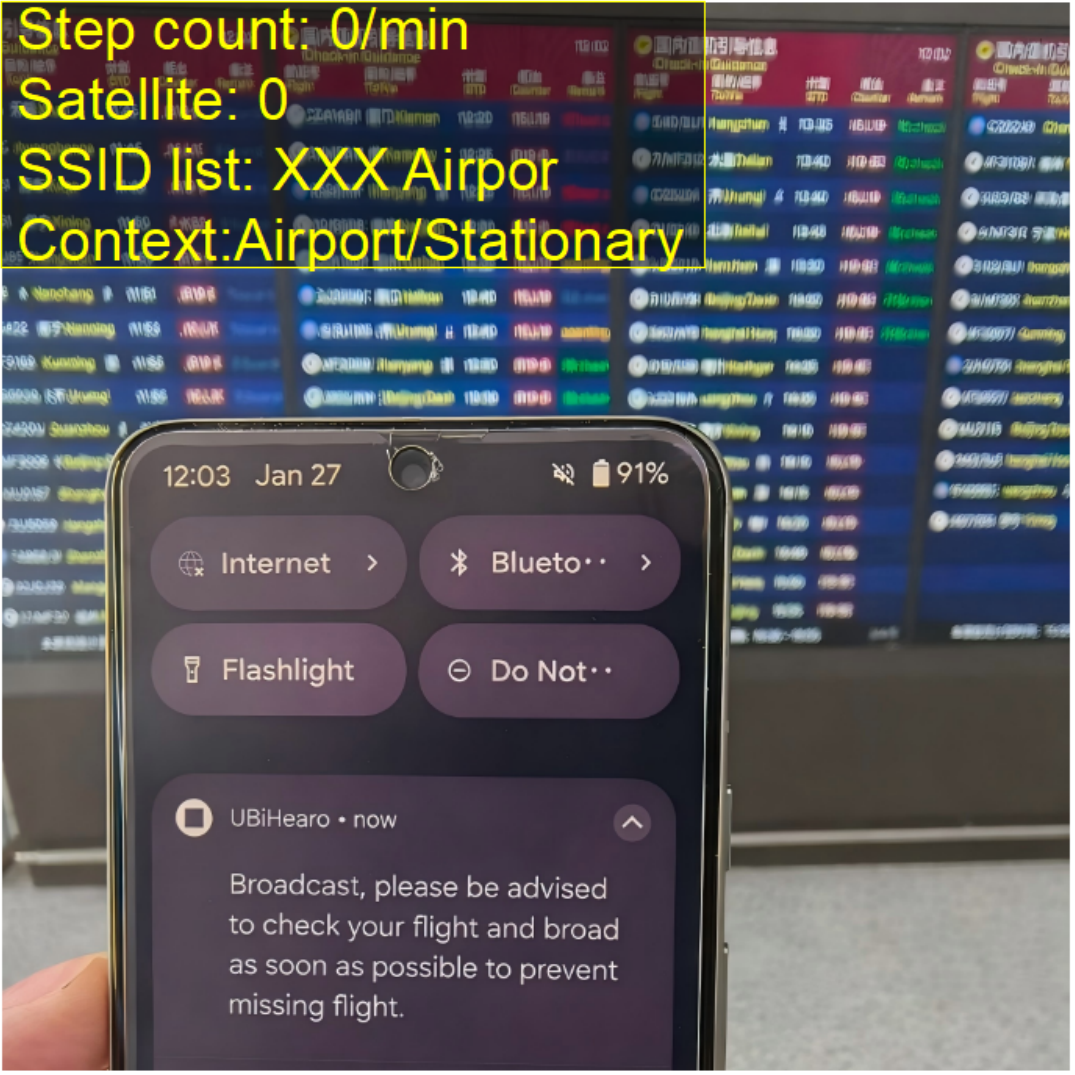}}
    \vspace{-1mm}
    \subfloat[Road]{\label{fig:road}
    \includegraphics[width=0.2\linewidth]{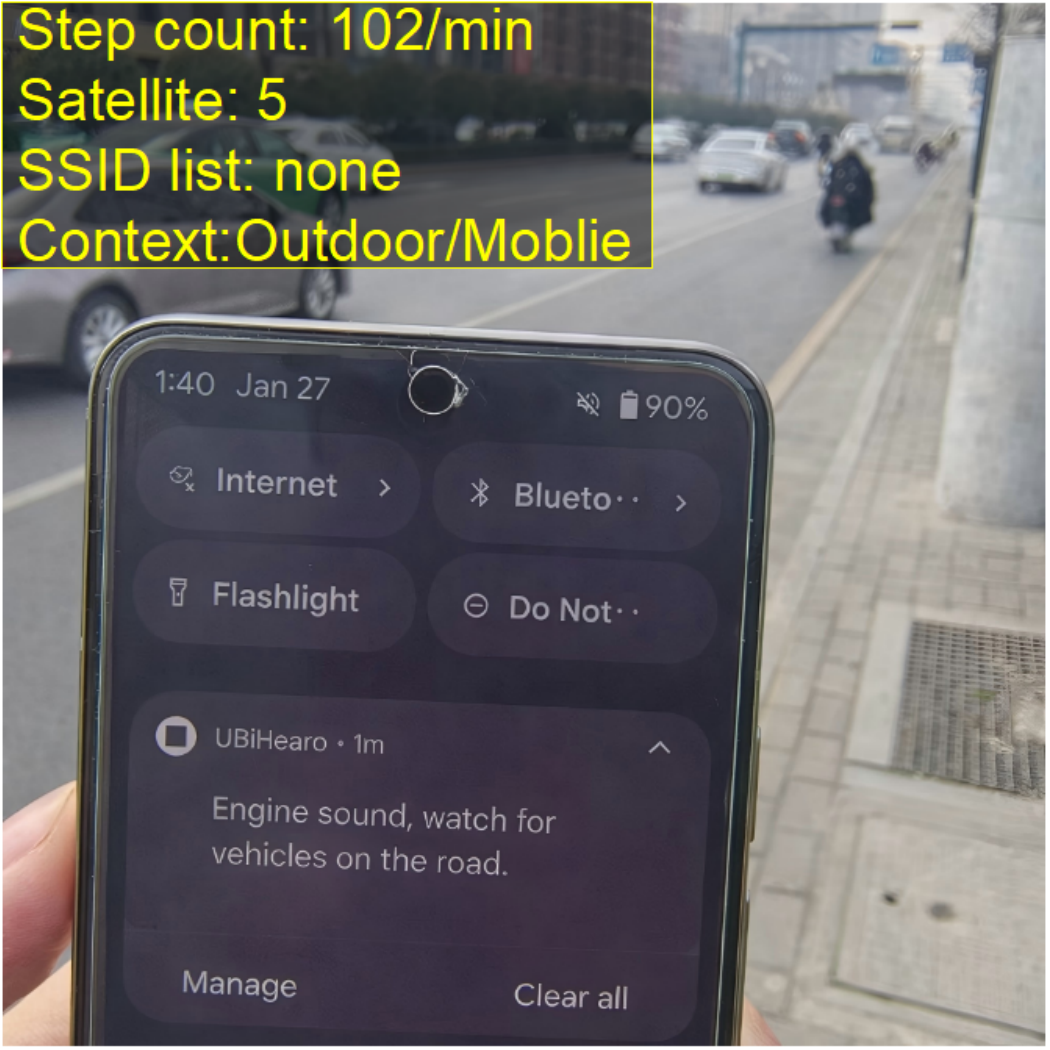}}
    \subfloat[Mall]{\label{fig:mall}
    \includegraphics[width=0.2\linewidth]{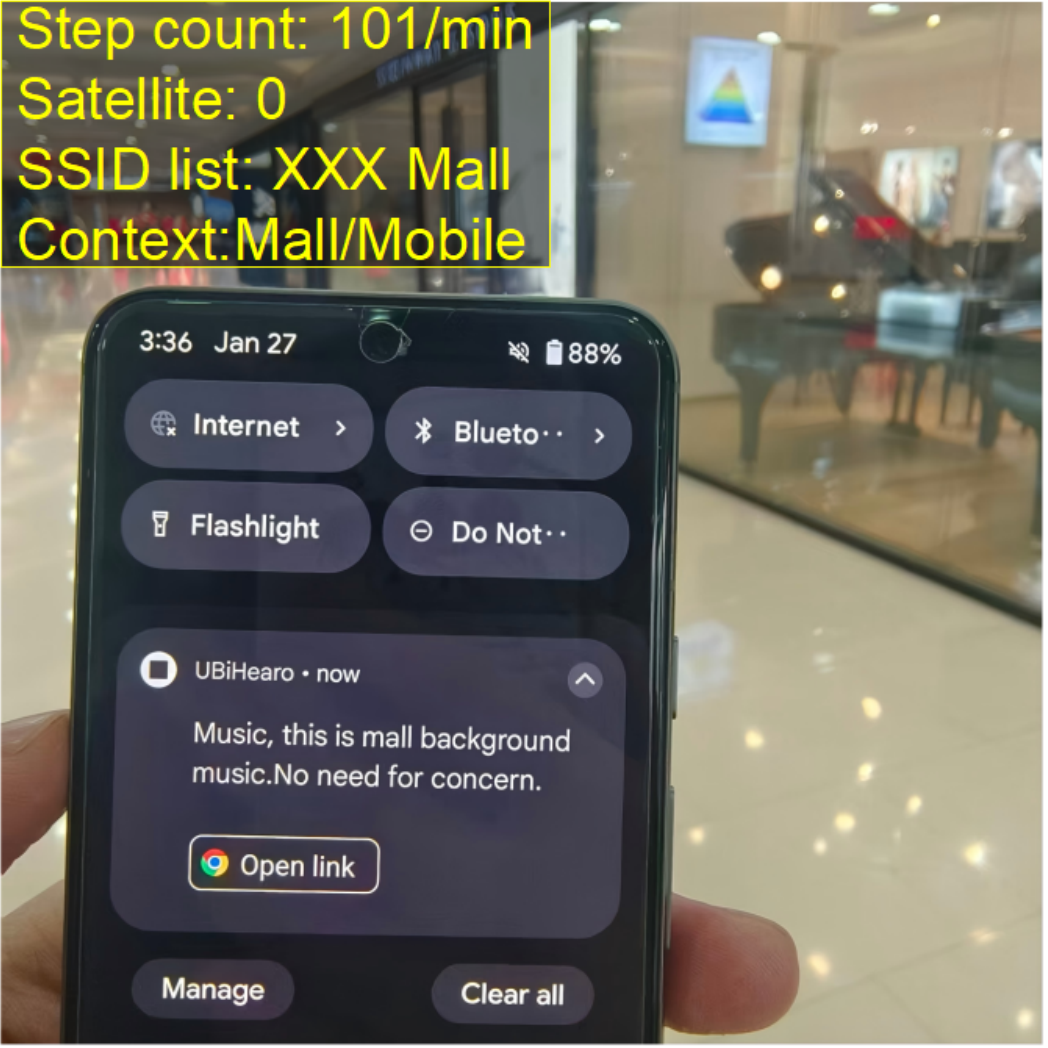}}
    \subfloat[Park]{\label{fig:water}
    \includegraphics[width=0.2\linewidth]{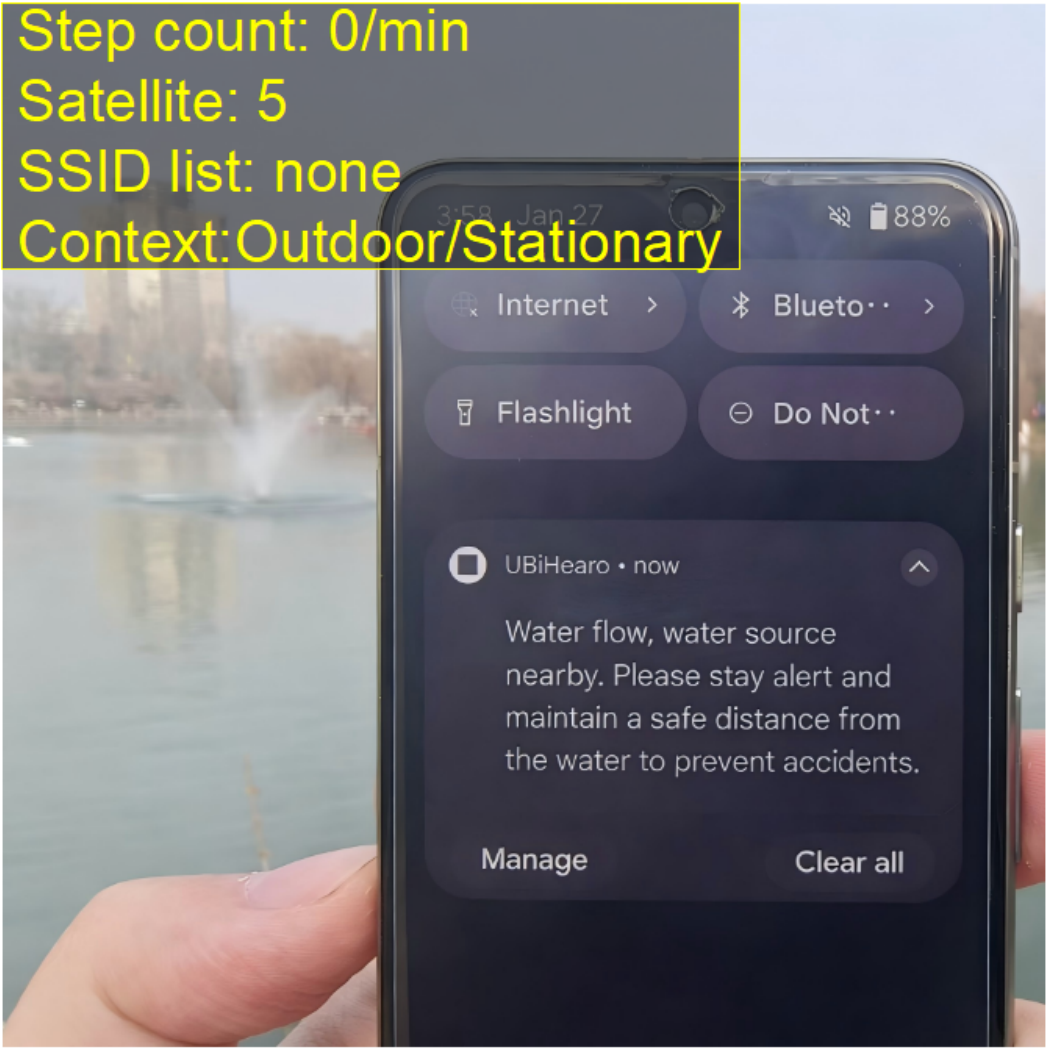}}
    \subfloat[In-vehicle]{\label{fig:car}
    \includegraphics[width=0.2\linewidth]{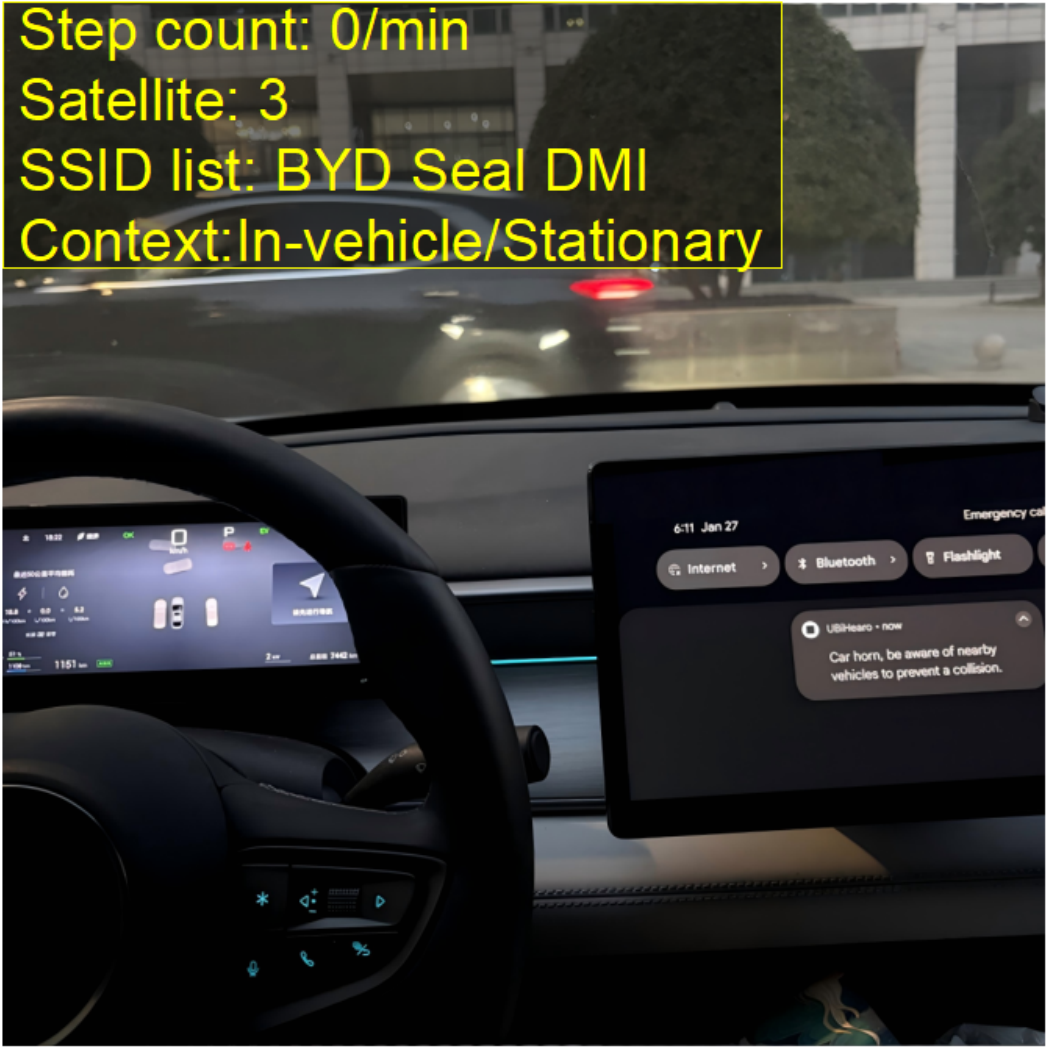}}
    \caption{In-the-wild case study on Pixel~8 across eight everyday scenarios, spanning diverse mobility, noise, and acoustic conditions (\eg interference, broadcasts, alarms, and horns) to assess timely, scenario-aware, actionable guidance.}
    \vspace{-5mm}
    \label{fig:case}
\end{figure}

\subsection{Case Study}
To examine \sysname in real-world conditions, we deploy it on a Pixel~8 across everyday environments, including railway stations, subways, airports, roadside traffic, shopping malls, parks, and in-vehicle settings.
These scenarios vary in mobility, background noise, interference, and salient sounds such as broadcasts, alarms, and horns.
As shown in \figref{fig:case}, we log representative on-device responses to assess whether the guidance is timely, scenario-aware, and actionable.

\textbf{In-the-moment examples.}
\sysname provides context-aware guidance under noisy and time-critical conditions.
On a train platform, it maps boarding broadcasts to time-sensitive advice: \textit{Broadcast, please note the train time and do not miss the boarding time} (\figref{fig:rail}).
In a crowded subway, a doors-close warning triggers safety guidance: \textit{Alert sound, please assess your surroundings and keep a safe distance from the source} (\figref{fig:alert}).
At an airport, thunder and boarding reminders lead to flight-aware suggestions (\figref{fig:air1}, \figref{fig:air2}).
Near dense traffic, engine noise triggers: \textit{Engine sound, watch for vehicles on the road} (\figref{fig:road}).
In benign continuous-sound settings, such as mall music and park fountains, \sysname correctly de-escalates notifications (\figref{fig:mall}, \figref{fig:water}).
Finally, when preparing to pull out of a parking spot and a nearby vehicle honks, \sysname issues a collision-aware warning (\figref{fig:car}).

\begin{figure}[t]
    \centering
    \vspace{-5mm}
    \subfloat[Power consumption]{\label{fig:c1}    \includegraphics[height=0.21\linewidth]{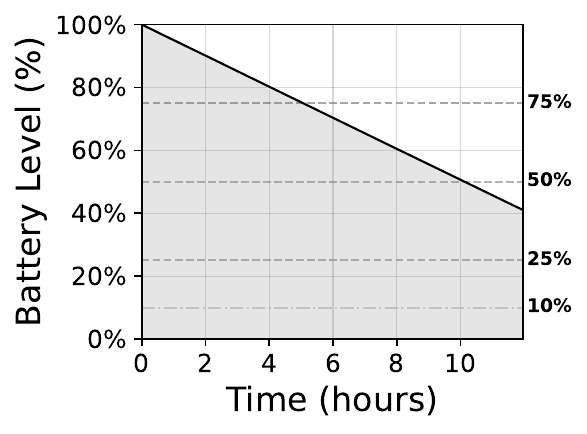}}
    \subfloat[System latency]{\label{fig:c2}    \includegraphics[height=0.21\linewidth]{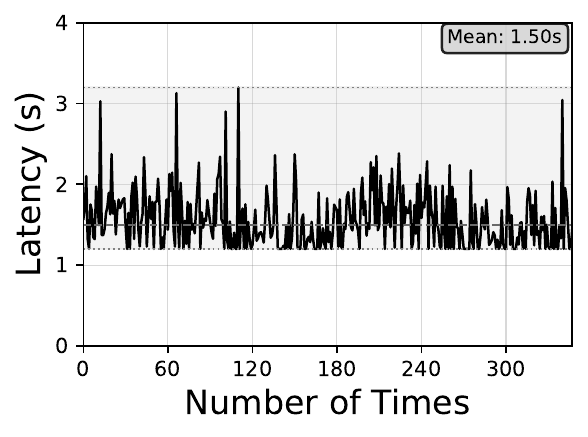}}
    \subfloat[Guidance quality (user-rated appropriateness)]{\label{fig:c3}    \includegraphics[height=0.21\linewidth]{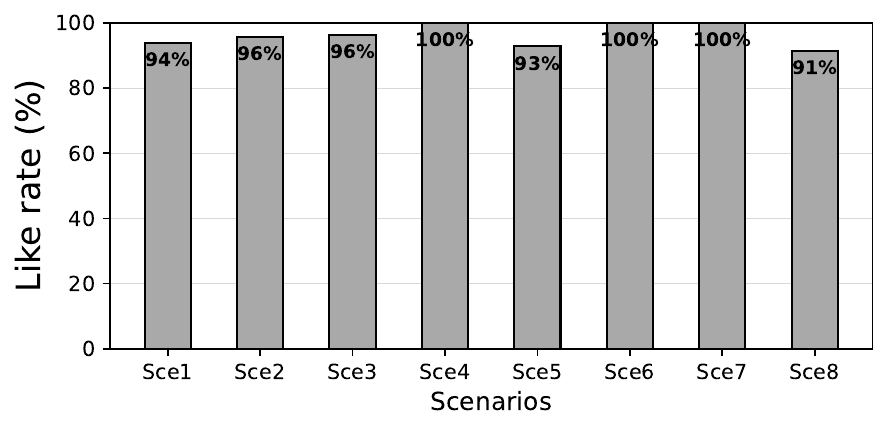}}
    \caption{Real-world system performance.}
    \vspace{-5mm}
    \label{fig:casep}
\end{figure}

\textbf{Real-world system performance.}
We further quantify continuous-use practicality with Pixel~8 running only the OS and \sysname, with the screen off except when triggered.
We log energy, latency, and output quality, and manually assess quality using the criteria in \secref{sec:DPO}: sound-type priorities, accuracy-latency trade-offs, and high-priority sounds.
Users mark appropriate outputs via an on-screen ``\textit{like}'' button (\figref{fig:diy}).
After $\sim$12 hours of continuous testing, the device retained 41\% battery (\figref{fig:c1}), and across 347 triggers, \sysname achieved 1.5\,s mean end-to-end latency (\figref{fig:c2}).

The 347 instances cover eight location labels: \textit{Sce1} Indoor (115), \textit{Sce2} Outdoor (73), \textit{Sce3} Railway station (28), \textit{Sce4} Airport (34), \textit{Sce5} Mall (14), \textit{Sce6} Restaurant (20), \textit{Sce7} Hospital (16), and \textit{Sce8} In-vehicle (47).
As shown in \figref{fig:c3}, user-rated appropriateness remains high across scenarios: 93.9\%, 95.8\%, 96.4\%, 100\%, 92.8\%, 100\%, 100\%, and 91.4\%, respectively.
Failures are explainable and concentrated in a few cases, such as weak satellite/WiFi signals limiting broadcast-to-arrival association, traffic-light \textit{ding-ding} sounds misread as alerts, train whistles confused with wind, video-wall audio misclassified as speech, and door slams during vehicle entry/exit not reliably recognized.
Speech-dominant environments, such as restaurants and hospitals, also bias detections toward speech, although most outputs still satisfy our criteria.

%% file: body/rel.tex
\section{RELATED WORK}

\subsection{Assistive Technologies and Human Augmentation for DHH Users}
Mobile human augmentation aims to restore perceptual access through lightweight, real-time assistance.
Prior DHH systems have explored AR/VR spatial visualization~\cite{bib:WeavingSound,bib:SonoCraftAR}, wearable haptics~\cite{bib:Haptic-captioning,bib:TactileEmotions}, and IoT-based environmental indicators~\cite{bib:Exploringsound}.
However, these solutions may suffer from hardware bulk, tactile fatigue, or limited portability.
In contrast, \sysname uses ubiquitous smartphones to support a hardware-free sensing--reasoning--action loop, moving from basic sound awareness to scenario-aware guidance.

\subsection{Transitioning Mobile Sound Awareness from Classification to Reasoning}
Mobile sound awareness has evolved from handcrafted features and CNN classifiers~\cite{bib:Environmental,bib:Optimization} to self-attention encoders~\cite{bib:Ast} and large pretrained audio models~\cite{bib:whisper}.
Although these methods improve recognition, they mostly produce closed-set labels and can degrade in noisy, dynamic settings~\cite{bib:SoundWatch,bib:Protosound}.
Recent personalization methods further support on-device adaptation~\cite{bib:AdaptiveSound,bib:SPECTRA}, but still focus mainly on recognition.
LLMs offer a shift from fixed labels to open-set reasoning~\cite{bib:dnnagent}.
\sysname builds on this shift by grounding acoustic tags with physical context, enabling scenario-aware guidance rather than ambiguous sound notifications.

\subsection{Human Preference Alignment Techniques}
Preference alignment is important for safety-related assistance, where outputs must reflect user priorities and acceptable actions.
RLHF~\cite{bib:Asurveyofreinforcement} is widely used but introduces reward modeling and online optimization costs that are unsuitable for lightweight mobile deployment.
DPO~\cite{bib:Directpreference} provides a more efficient alternative by directly optimizing preference likelihoods.
\sysname adopts DPO to align generative guidance with DHH users' safety priorities and lifestyle preferences while keeping adaptation lightweight.

%% file: body/con.tex
\section{CONCLUSION}
This paper addresses a key unmet need in mobile hearing assistance for Deaf and Hard-of-Hearing (DHH) users: moving beyond recognition-only sound labels toward \textit{always-on, scenario-aware} understanding and \textit{actionable guidance}.
We present \sysname, a mobile sound guidance agent that closes the \textit{sensing--reasoning--action} loop through energy-efficient duty-cycled sensing, structured acoustic/context representations, staged on-device reasoning, and lightweight human-in-the-loop personalization.
By automatically generating structured prompts from acoustic tags and physical context, \sysname reduces the burden of user-crafted prompts and enables scenario-aware guidance in everyday environments.
Across mobile devices and real-world tasks, \sysname improves guidance quality while maintaining practical energy and latency costs.
Future work will incorporate richer low-power context signals and explore more efficient alignment and distillation methods to further reduce runtime cost while preserving user-aligned guidance.

%% file: body/appendix.tex
\section{APPENDIX}

In this appendix, we present the questionnaire used in the survey study described in \secref{sec:background}, along with the algorithm for the coarse-grained acoustic event detector introduced in \secref{sec:4.1}.
\subsection{Questionnaire}
\label{questionnaire}
\begin{enumerate}
\item Which channels do you usually use to obtain sound information? (Single choice) \\
Cochlear implant:
\\~~$\Box$~Always~~$\Box$~Sometimes~~$\Box$~Occasionally~~$\Box$~Rarely~~$\Box$~Never\\
Hearing aid:
\\~~$\Box$~Always~~$\Box$~Sometimes~~$\Box$~Occasionally~~$\Box$~Rarely~~$\Box$~Never\\
Someone informs you via sign language:
\\~~$\Box$~Always~~$\Box$~Sometimes~~$\Box$~Occasionally~~$\Box$~Rarely~~$\Box$~Never\\
Ask friends or family:
\\~~$\Box$~Always~~$\Box$~Sometimes~~$\Box$~Occasionally~~$\Box$~Rarely~~$\Box$~Never\\
Use a phone app:
\\~~$\Box$~Always~~$\Box$~Sometimes~~$\Box$~Occasionally~~$\Box$~Rarely~~$\Box$~Never\\
Use a watch app:
\\~~$\Box$~Always~~$\Box$~Sometimes~~$\Box$~Occasionally~~$\Box$~Rarely~~$\Box$~Never\\

\item How often do you miss sounds you are interested in? (Single choice) \\
At home:\\
~~$\Box$~More than once/day~~$\Box$~Once/day~~$\Box$~Once/week~~$\Box$~Once/month~~$\Box$~None\\
At work/school:\\
~~$\Box$~More than once/day~~$\Box$~Once/day~~$\Box$~Once/week~~$\Box$~Once/month~~$\Box$~None\\
When you are moving and looking at your phone:\\~~$\Box$~More than once/day~~$\Box$~Once/day~~$\Box$~Once/week~~$\Box$~Once/month~~$\Box$~None\\
When you are moving but not looking at your phone:\\~~$\Box$~More than once/day~~$\Box$~Once/day~~$\Box$~Once/week~~$\Box$~Once/month~~$\Box$~None\\

\item Preference differences across scenarios (Single choice for each item) \\
Scenario A: Indoors and stationary\\
In this scenario, which factor do you prioritize the most for a sound-awareness assistant? (Single choice)\\
~~$\Box$~Recognition accuracy (whether the sound is correctly recognized)\\
~~$\Box$~Response latency (how quickly the prompt is delivered)\\
~~$\Box$~Sound category coverage (whether it can recognize the sounds I care about)\\
~~$\Box$~Other:\\
Scenario B: Outdoors and mobile\\
~~$\Box$~In this scenario, which factor do you prioritize the most for a sound-awareness assistant? (Single choice)\\
~~$\Box$~Recognition accuracy (whether the sound is correctly recognized)\\
~~$\Box$~Response latency (how quickly the prompt is delivered)\\
~~$\Box$~Sound category coverage (whether it can recognize the sounds I care about)\\
~~$\Box$~Other:\\
In addition, in these two scenarios, which type of sounds do you care about the most? (Single choice for each item)\\
Scenario A: Indoors and stationary (Single choice)\\
~~$\Box$~Human speech and socially meaningful sounds (\eg conversations, someone calling my name)\\
~~$\Box$~Safety-critical sounds (\eg alarms, gas leak alerts)\\
~~$\Box$~Home environment sounds (\eg, doorbell, knocking, appliance sounds)\\
~~$\Box$~Other:\\
Scenario B: Outdoors and mobile (Single choice)\\
~~$\Box$~Safety-critical ambient sounds (\eg traffic sounds, horns, approaching vehicles)\\
~~$\Box$~Human speech and socially meaningful sounds (\eg people calling, public announcements)\\
~~$\Box$~Natural environment sounds (\eg wind, rain, thunder)\\
~~$\Box$~Other:\\
Finally, do you think your preferences for the assistant (\eg accuracy vs. speed, sound categories) will change over time due to daily routines, occupation, or familiarity with the assistant? (Single choice)\\
~~$\Box$~Yes, it will change frequently\\
~~$\Box$~Yes, it will change sometimes\\
~~$\Box$~Not sure\\
~~$\Box$~No, it will not change\\
If yes, what is the main reason? (Optional): \\

\item If the same sound is given the same prompt without distinguishing scenarios, would you feel that it adds hassle to your life or poses a danger? \\
(1):When you hear “a horn honking”:\\
A. Indoors (should remind: Can be ignored)\\
B. On the road (should remind: Watch out for vehicles)\\
If scenarios are not distinguished, and both scenarios prompt you “Can be ignored”, what do you think? (Single choice)\\
~~$\Box$~Dangerous~~$\Box$~Adds hassle~~$\Box$~Minor impact~~$\Box$~No impact\\
(2):When you hear “screaming”:\\
A. Amusement park (should remind: Someone nearby is playing)\\
B. Outdoors (should remind: Look for the source of the scream)\\
If scenarios are not distinguished, and both scenarios prompt you “Someone nearby is playing”, what do you think? (Single choice)\\
~~$\Box$~Dangerous~~$\Box$~Adds hassle~~$\Box$~Minor impact~~$\Box$~No impact\\
(3): When you hear “a broadcast”:\\
A. Music broadcast in the square (should remind: Can be ignored)\\
B. High-speed rail station/airport (should remind: Pay attention to the departure time of the train/flight)\\
If scenarios are not distinguished, and both scenarios prompt you “Can be ignored”, what do you think? (Single choice)\\
~~$\Box$~Dangerous~~$\Box$~Adds hassle~~$\Box$~Minor impact~~$\Box$~No impact\\
(4): When you hear “running water”:\\
A. At home (should remind: Please turn off the tap)\\
B. In the wild (should remind: Watch out for nearby hazardous water sources)\\ 
If scenarios are not distinguished, and both scenarios prompt you “Please turn off the tap”, what do you think? (Single choice)\\
~~$\Box$~Dangerous~~$\Box$~Adds hassle~~$\Box$~Minor impact~~$\Box$~No impact\\
(5): When you hear “thunder”:\\
A. Indoors (should remind: Can be ignored)\\
B. Outdoors (should remind: Watch out for lightning)\\
If scenarios are not distinguished, and both scenarios prompt you “Can be ignored”, what do you think? (Single choice)\\
~~$\Box$~Dangerous~~$\Box$~Adds hassle~~$\Box$~Minor impact~~$\Box$~No impact\\

\item Scenario description: \\
Assume the phone recognizes a sound and will give you a prompt. Of the following two prompts, which do you think is more useful?\\
Prompt 1 (only tells what sound it is): “An alarm is sounding”\\
Prompt 2 (tells what sound it is + tells you what to do): “An alarm is sounding; suggestion: first check the kitchen/living room; if you smell gas, open the window, shut off the valve, leave, and seek help.”\\
(1):Which prompt would you prefer? (Single choice)\\
~~$\Box$~Prompt 1: only tells the sound\\
~~$\Box$~Prompt 2: sound + suggestion\\
~~$\Box$~Either is fine\\
~~$\Box$~I don’t want either\\
(2):What is the main reason you chose it? (Multiple choice, select up to 2)\\
~~$\Box$~I want to know what to do next\\
~~$\Box$~Only seeing the “sound name”, I’m still not sure whether it is serious / whether I should handle it immediately\\
~~$\Box$~Having suggestions makes me feel more reassured / safer\\
~~$\Box$~Only stating the sound is enough; I can judge by myself\\
~~$\Box$~Too many prompts would feel annoying / disruptive\\
~~$\Box$~Other: \\
(3):If the phone only prompts the sound (for example, “An alarm is sounding”), would you immediately know what to do? (Single choice)\\
~~$\Box$~Yes, I would immediately know what to do\\
~~$\Box$~Sometimes; it depends on the situation\\
~~$\Box$~Often not; I need to think about it / ask others\\
~~$\Box$~No; I don’t know what this means\\

\end{enumerate}
\newpage
\subsection{Algorithm}
\label{algorithm}
\begin{algorithm}[!h]
\caption{Human Speech / Ambient Sound / Safety-Critical Detection Algorithm}
\label{alg:voice_env_safety}
\begin{algorithmic}[1]
\Require 
    \parbox[t]{0.8\textwidth}{
        $Detector_{result}$: Detect single-frame decision (Silence / Non-silence)\\
        $E_k$: Energy of each band ($k\in \{0,1,...,6\}$)\\
        $T_E(k)$: Energy threshold of band $k$\\
        $R_k$: Energy ratio of band $k$ (\%)\\
        $E_{min}$: Silence threshold\\
        $\Delta_t$: Temporal window for consistency check
    }
\Ensure 
    $Final_{result}$: (0=Silence, 1=Speech, 2=Ambient, 3=Safety-critical)

\State $E_{total} \leftarrow \sum_k E_k$

\If{$E_{total} \leq E_{min}$}
    \State $Final_{result} \leftarrow 0$
\Else
    \If{$Detector_{result} = \text{Non-silence}$}

        \Comment{\textbf{Step 1: Safety-critical pattern detection}}

        \State $Siren_{pattern} \leftarrow (R_{400\sim2000} \geq 35) \land (TemporalOscillation(\Delta_t) = True)$

        \State $Alarm_{pattern} \leftarrow (R_{2\sim4kHz} \geq 30) \land (PeakRepeat(\Delta_t) = True)$

        \If{$Siren_{pattern} \lor Alarm_{pattern}$}
            \State $Final_{result} \leftarrow 3$ 
            \Comment{Safety-critical event → deterministic override}
        \Else

            \Comment{\textbf{Step 2: Human speech verification}}

            \State $Core_1 \leftarrow (E_{400\sim800} \geq T_E) \land (R_{400\sim800} \geq 20)$
            \State $Core_2 \leftarrow (E_{1600\sim3200} \geq T_E) \land (R_{1600\sim3200} \geq 15)$
            \State $Core_{ok} \leftarrow Core_1 \land Core_2$

            \Comment{\textbf{Step 3: Noise filtering}}

            \State $Noise_1 \leftarrow (R_{0\sim200} \leq 10)$
            \State $Noise_2 \leftarrow (R_{4\sim16kHz} \leq 8)$
            \State $Noise_{ok} \leftarrow Noise_1 \land Noise_2$

            \Comment{\textbf{Step 4: Auxiliary conditions}}

            \State $Aux_1 \leftarrow (R_{200\sim400} \geq 5)$
            \State $Aux_2 \leftarrow (R_{800\sim1600} \geq 8)$
            \State $Aux_{ok} \leftarrow Aux_1 \lor Aux_2$

            \If{$Core_{ok} \land Noise_{ok} \land Aux_{ok}$}
                \State $Final_{result} \leftarrow 1$
            \Else
                \State $Final_{result} \leftarrow 2$
            \EndIf
        \EndIf

    \Else
        \State $Final_{result} \leftarrow 2$
    \EndIf
\EndIf

\Return $Final_{result}$
\end{algorithmic}
\end{algorithm}